\documentclass{article}

\PassOptionsToPackage{numbers}{natbib}

 \usepackage[main, final]{neurips_2026}

\usepackage[utf8]{inputenc} 
\usepackage[T1]{fontenc}    
\usepackage{hyperref}       
\usepackage{url}            
\usepackage{booktabs}       
\usepackage{amsfonts}       
\usepackage{nicefrac}       
\usepackage{microtype}      
\usepackage{xcolor}         
\usepackage{graphicx}
\usepackage[ruled,vlined]{algorithm2e}
\usepackage{booktabs}
\usepackage{tabularx}
\usepackage{array}
\usepackage{threeparttable}
\usepackage{wrapfig}
\usepackage{caption}
\usepackage{amsmath}
\usepackage{float}
\usepackage{makecell}

\title{ElasticFit: Fit-Aware 3D Object Insertion via VLM Reasoning and Generative Adaptation}

\author{%
  Tzu-Hsin Hsieh \\
  Delft University of Technology \\
  Delft, Netherlands \\
  \texttt{thhsieh@tudelft.nl} \\
  \And
  Ricardo Marroquim \\
  Delft University of Technology \\
  Delft, Netherlands \\
  \texttt{R.Marroquim@tudelft.nl} \\
}

\begin{document}

\maketitle
\vspace{-3.0em}
\begin{center}
    \includegraphics[width=\linewidth]{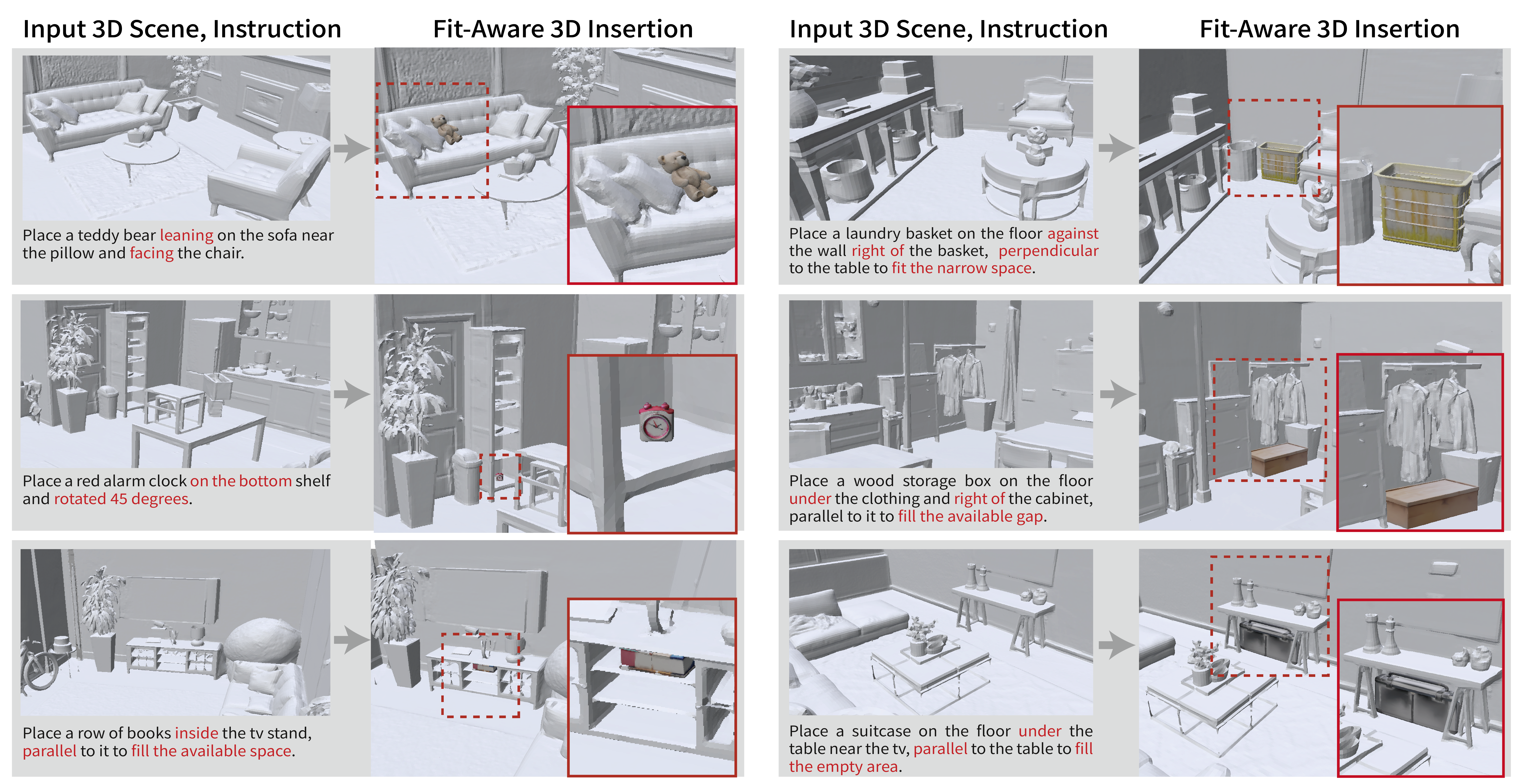}
    \captionof{figure}{
    \textbf{ElasticFit} inserts generated 3D objects into pre-existing scenes from natural-language instructions. It combines VLM reasoning, generative adaptation, and geometry- and physics-aware refinement. Red text highlights key constraints; results are shown in color.
    }
    \label{fig:teaser}
\end{center}

\begin{abstract}
    Inserting objects into existing 3D scenes requires more than selecting a plausible location:
    the inserted object must also fit local geometry while preserving semantic intent and physical plausibility.
    Although recent Vision-Language Models (VLMs) and generative models enable semantic reasoning and visual content creation, they offer limited 3D grounding and geometric control when an inserted object must fit into constrained local spaces.
    We introduce \textbf{ElasticFit}, a VLM-guided framework for fit-aware object insertion centered on a novel scene-grounded representation.
    Given a language instruction and rendered scene observations, ElasticFit infers structured fitting cues that specify where the object should be grounded, what volume it should occupy, how it should be oriented, and its adaptation mode (rigid placement, uniform scaling, or elastic fitting).
    These cues convert high-level VLM reasoning into explicit 3D constraints that condition object generation and guide downstream geometric fitting.
    ElasticFit then generates a scene-conditioned object prior, reconstructs it in 3D, and refines the mesh through mode-specific fitting while enforcing collision avoidance, contact consistency, and physical grounding.
    In fixed-asset baseline comparisons, ElasticFit improves spatial relation success from 50.8\% to 69.7\% and support success from 48.3\% to 91.7\% over the strongest baseline, while providing novel support for generative "make-it-fit" insertions in complex scenarios.
    Project page: \url{https://celine-hsieh.github.io/elasticfit/}.
\end{abstract}

\section{Introduction}

Inserting objects into existing 3D scenes is a fundamental capability for augmented and virtual reality, automated interior design, and embodied AI \cite{hellmuth2021interaction,abdelreheem2025placeit3d,paschalidou2021atiss,yang2024holodeck}. Unlike isolated object generation or holistic scene synthesis, this task requires reasoning about an already occupied environment where a newly inserted object must satisfy multiple constraints. A successful insertion must not only pinpoint a location but also align with the user’s language intent and fit the local geometry \cite{huang2025fireplace}. Furthermore, it must account for available free space, support surfaces, contact, and collision constraints. 
This makes object insertion a joint reasoning problem over semantic intent, scene context, and physical feasibility.

Recent progress in 3D placement and layout generation has demonstrated the potential of Vision-Language Models (VLMs) \cite{openai2023gpt4v, team2023gemini} for high-level scene understanding and commonsense reasoning \cite{fu2025scene, feng2023layoutgpt, sun2025layoutvlm, yang2024holodeck}. 
These methods can often infer plausible object categories and coarse placement regions from scene context and user intent. 
However, most existing approaches primarily focus on selecting, retrieving, or positioning a given object, and remain limited when successful insertion requires the object itself to adapt to the local geometry of the target scene. 
In such cases, the challenge lies not only in deciding \emph{where} to place an object, but also in determining \emph{how} its shape, scale, and pose should adapt to fit constrained space while remaining semantically appropriate and physically feasible. 
As a result, existing methods are typically restricted to pose adjustment or global scaling, rather than adapting object geometry to local free space \cite{tang2024diffuscene, fu20213d}. 
This limitation becomes especially critical in cluttered or spatially constrained scenes.

These limitations suggest three key capabilities required for fit-aware insertion.
\textbf{Geometry-grounded scene understanding.}
Successful insertion requires reasoning over actual support surfaces and surrounding free space, rather than relying only on coarse abstractions such as bounding boxes \cite{huang2024blenderalchemy, aguina2024open,hu2024scenecraft,paschalidou2021atiss, yang2024holodeck}. 
ElasticFit grounds insertion in local scene geometry through scene-grounded target regions and fitting constraints.
\textbf{Make-it-fit adaptation.}
Most existing methods assume that a suitable object is already available, either retrieved from an asset repository \cite{tang2024diffuscene, yang2024holodeck, ocal2024sceneteller} or provided for placement optimization \cite{huang2025fireplace, abdelreheem2025placeit3d}. 
This assumption fails when the target region requires the object itself to adapt to surrounding spatial constraints. 
ElasticFit treats insertion as constrained fitting, enabling rigid alignment, uniform scaling, or elastic adaptation.
\textbf{Physics-aware refinement.}
Although prior methods often enforce geometric constraints, their final refinement is typically not designed to settle inserted objects under physical constraints \cite{huang2025fireplace, yang2024holodeck, sun2025layoutvlm, paschalidou2021atiss}. 
As a result, placements may still suffer from intersections, weak support, floating artifacts, or overhangs. 
ElasticFit refines the generated object with geometry- and physics-aware fitting objectives to improve contact, support, and collision compatibility.

We propose \textbf{ElasticFit}, a method for fit-aware 3D object insertion that converts language instructions and rendered scene observations into scene-grounded fitting constraints. Rather than directly predicting a final object transform, ElasticFit infers a scene-grounded fitting representation $G$, including the target region, orientation, support and contact requirements, and adaptation mode. This representation conditions object generation and mode-specific deformation-aware fitting, allowing the inserted object to be rigidly aligned, uniformly scaled, or elastically adapted to local spatial constraints. By coupling semantic grounding with geometry- and physics-aware optimization, ElasticFit produces insertions that are contextually appropriate, spatially aligned, and physically plausible.

Our contributions are summarized as follows:
\begin{itemize}
    \item \textbf{Fit-aware object insertion formulation:} 
    We formulate object insertion into pre-existing 3D scenes as fit-aware adaptation, jointly considering semantic intent, local geometric compatibility, and object-shape adaptation.
    
    \item \textbf{Scene-grounded fitting and adaptation:} 
    We introduce a structured representation $G$ that converts VLM-inferred cues into grounded constraints for object generation and rigid, uniform, or elastic fitting.

    \item \textbf{Experiments:} 
    We evaluate ElasticFit on two task sets covering common placement relations and fit-critical constrained insertions. 
    In fixed-asset baseline comparisons, ElasticFit improves spatial relation success from 50.8\% to 69.7\% and support success from 48.3\% to 91.7\% over the strongest baseline, and further demonstrates fit-aware insertion under pose, contact, and constrained-space requirements.
\end{itemize}

\section{Related Work}
\label{related_work}
\textbf{Placement and Editing in Existing 3D Scenes.}
Recent work has explored language-guided object placement and editing in pre-existing 3D scenes, including placement in complex asset-based environments, placement in real scanned scenes, and language-guided scan or layout editing~\cite{huang2025fireplace,abdelreheem2025placeit3d,el2025scanedit,zheng2024editroom}. 
FirePlace~\cite{huang2025fireplace} is the closest prior work, as it grounds MLLM reasoning in fine-grained scene geometry and solves geometric constraints for placing a given 3D object in a pre-existing scene. 
PLACEIT3D~\cite{abdelreheem2025placeit3d} formulates language-guided object placement in real 3D scenes and introduces a benchmark for this task. 
ScanEdit~\cite{el2025scanedit} and EditRoom~\cite{zheng2024editroom} address instruction-driven editing of existing 3D scans or room layouts through hierarchical reasoning or layout-level editing. 
Despite their differences, these methods primarily optimize where an object should be placed or how an existing scene should be rearranged, while the inserted asset is typically treated as fixed during placement or editing. 
ElasticFit studies a complementary make-it-fit setting: rather than only estimating the transformation of a fixed target asset, it generates and adapts the inserted object to a scene-grounded fitting region. 
This distinction is important in constrained local spaces, such as narrow gaps, shelves, or regions under furniture, where the object geometry itself may need to change to satisfy local spatial affordances. We provide a detailed design-level comparison with FirePlace in Appendix~\ref{app:comparison}.

\textbf{Foundation-Model-Guided 3D Layout Planning.}
A broad line of work studies 3D layout generation and scene synthesis using learned priors, structured scene representations, and optimization over object arrangements~\cite{paschalidou2021atiss,wang2021sceneformer,tang2024diffuscene,lin2024instructscene,wang2019planit,zhai2023commonscenes,yang2024physcene}. 
Recent foundation-model-guided methods further leverage language or vision-language reasoning to improve semantic planning and controllability~\cite{sun20253d,zhou2025scenex,ocal2024sceneteller,ccelen2024idesign,hu2024scenecraft,tam2024scenemotifcoder,ling2025scenethesis}. 
Representative examples include LayoutGPT~\cite{feng2023layoutgpt}, which predicts layouts from textual conditions; Holodeck~\cite{yang2024holodeck}, which generates spatial constraints and solves them through search-based construction; and LayoutVLM~\cite{sun2025layoutvlm}, which performs visually guided layout reasoning and optimization from multi-view inputs. 
These methods are closely related to ElasticFit's semantic-spatial reasoning and form the main baseline family. 
However, they mainly operate on layout abstractions such as object poses, bounding boxes, or symbolic constraints, which are insufficient for fine-grained support, contact, and make-it-fit adaptation in complex pre-existing scenes. 
In contrast, ElasticFit couples semantic-spatial reasoning with geometry-aware adaptation, enabling inserted objects to conform to local scene structure.

\textbf{Generative Adaptation and Asset Creation.}
Recent advances in generative models have led to highly capable text-to-image models~\cite{podell2023sdxl, lee2024koala, rombach2022high}, which in turn have enabled strong 2D-to-3D and text-to-3D asset creation pipelines~\cite{xiang2025structured, xu2024instantmesh, zhao2025hunyuan3d, tochilkin2024triposr}. 
Related work has also begun to explore generative object insertion in existing 3D representations, including neural radiance fields and 3D Gaussian scenes~\cite{shahbazi2024inserf, li2025freeinsert, madhavaram2025towards, bartrum2024replaceanything3d, xiao2025localized}. 
These methods demonstrate that generation can be used not only to create assets in isolation, but also to introduce new objects into scene context. 
However, they mainly formulate the problem as generative insertion or editing, rather than as scene-grounded adaptation for satisfying local geometric constraints. 
ElasticFit instead uses generation as part of a fit-aware insertion process, where the inserted asset is shaped jointly by semantic intent, local geometry, and contextual appearance.

\begin{figure*}[!t]
  \centering
  \includegraphics[width=\textwidth]{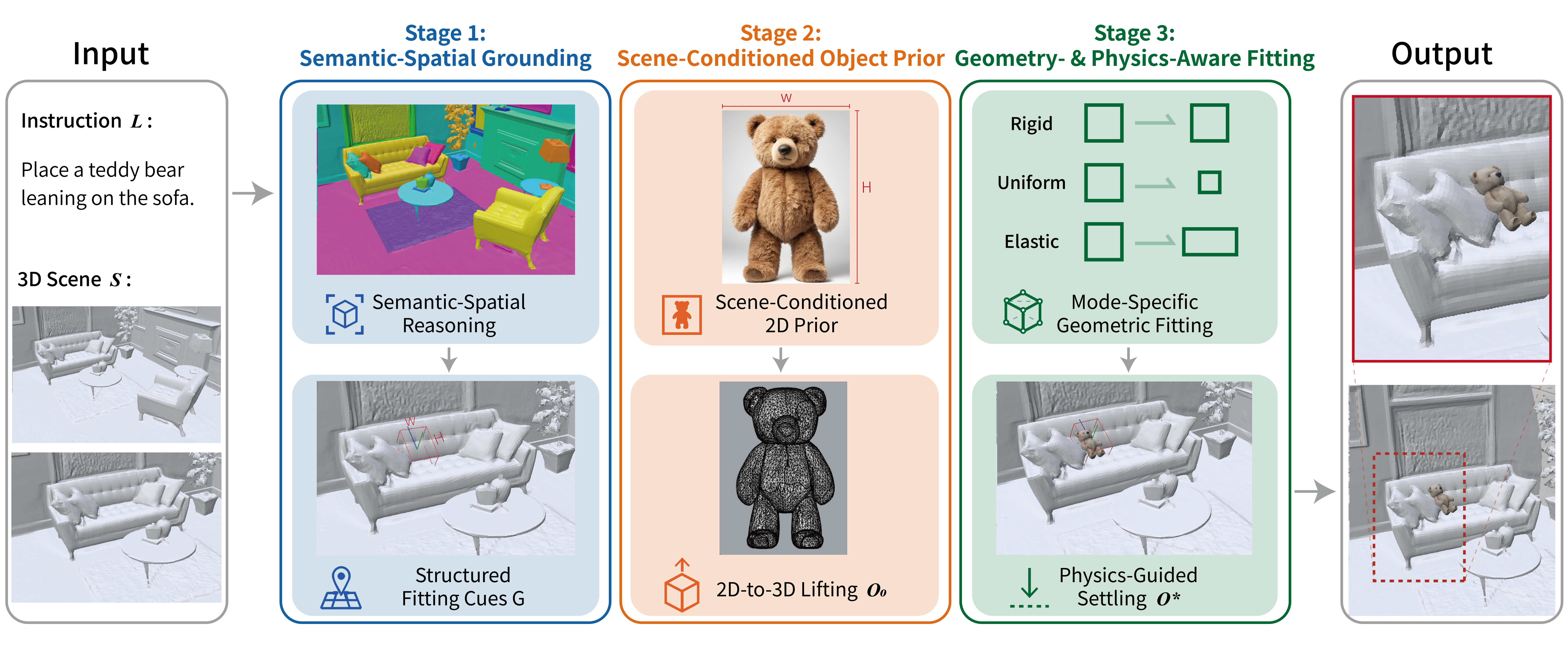}
  \caption{
    Overview of ElasticFit.
    Given a user instruction $L$ and an existing 3D scene $S$, ElasticFit infers structured fitting cues $G$ that encode the target region, orientation, constraints, and adaptation behavior.
    \textbf{Stage 1} performs semantic-spatial grounding,
    \textbf{Stage 2} generates a scene-conditioned object prior $O_0$.
    \textbf{Stage 3} refines it through geometry- and physics-aware fitting with rigid, uniform, or elastic adaptation. The result is a stable final insertion $O^\ast$.
  }
  \label{fig:pipeline}
\end{figure*}

\section{Method}
\label{method}




\subsection{Overview}


Given an existing 3D scene $S$ and a natural-language placement prompt $L$, ElasticFit inserts a new object so that the result is semantically appropriate, geometrically compatible, and physically plausible. 
Rather than treating placement as rigid pose prediction, ElasticFit formulates object insertion as fitting an object to local scene geometry under semantic and physical constraints. 
As illustrated in Figure~\ref{fig:pipeline}, ElasticFit realizes this formulation through three stages: 
\textbf{(1) Semantic-Spatial Grounding}, 
\textbf{(2) Scene-Conditioned Object Prior}, and 
\textbf{(3) Geometry- and Physics-Aware Fitting}.


The semantic-spatial grounding stage analyzes the scene and language input to infer a structured grounding representation $G$, formally defined in Sec.~\ref{stage1}. 
This representation encodes the geometric placement constraints, including the support point, ghost-box center and dimensions, and object orientation, as well as a deformation mode represented by an adaptation scale vector $\mathbf{s}\in\mathbb{R}_+^3$ and a semantic object description. 
Conditioned on $G$, the scene-conditioned object prior stage generates an initial object hypothesis $O_0$ by producing a scene-conditioned 2D object image and lifting it into a 3D mesh. 
The geometry- and physics-aware fitting stage then refines $O_0$ according to the geometric constraints and deformation behavior encoded in $G$, followed by physics-based settling to obtain a stable final insertion $O^\ast$. 
This design separates high-level semantic-spatial reasoning from downstream geometric realization, while connecting them through the unified grounding representation $G$.

\subsection{Semantic-Spatial Grounding [Stage 1]}
\label{stage1}
Given a scene $S$ and instruction $L$, the grounding stage maps VLM predictions into structured fitting cues $G$ that specify object position, orientation, support, and adaptation.

\vspace{-0.7em}
\paragraph{Semantic-Spatial Reasoning.}
We construct a view-aligned scene representation for VLM-based reasoning by rendering the scene from the same viewpoint shown to users. 
Using this view, scene geometry, and dataset-provided semantic-instance annotations, we build two geometric buffers: an XYZ buffer storing per-pixel 3D coordinates and an ID buffer storing the visible semantic instance ID at each pixel. 
Based on the ID buffer, we generate a Set-of-Mark (SoM)~\cite{yang2023set} image that overlays visible object IDs on the current view and summarizes candidate anchor objects. 
We then query a VLM with the unannotated rendered scene view, the SoM image, and the instruction to predict a structured JSON containing 
(i) a reference anchor object \texttt{anchor\_object\_id}, 
(ii) a target placement pixel \texttt{target\_uv} and support object \texttt{support\_object\_id}, 
(iii) optional orientation cues including \texttt{look\_at\_uv} and \texttt{top\_uv}, 
(iv) snapping behavior, and 
(v) when adaptive generation is enabled, object-generation fields, canonical object scale, adaptation tags, and additional fitting controls.

\vspace{-0.7em}
\paragraph{Structured Fitting Cues.}
We transform the VLM prediction from image-space cues into explicit 3D fitting conditions. 
We first validate the predicted \texttt{target\_uv} against the semantic ID buffer using \texttt{support\_object\_id}; if the predicted pixel does not lie on the intended support object, we replace it with the nearest visible pixel on that object. 
The corrected pixel is mapped through the XYZ buffer to obtain the 3D support point, which serves as the target surface point for grounding. 
Orientation is estimated from optional \texttt{look\_at\_uv} and \texttt{top\_uv} cues, with pose-specific priors, yielding yaw and pitch angles in world coordinates. 
Using the support point, predicted object dimensions, and inferred orientation, we construct a 3D ghost box whose bottom-face center is placed at the support point. 
The ghost box is not a tight object bounding box, but an estimated maximum fitting volume that guides downstream object generation and adaptation. Stage~1 outputs a structured grounding representation
\begin{equation}
\label{stage1_structure}
G = \{p_{\mathrm{surf}},\, c_{\mathrm{box}},\, \mathbf{b}_{\mathrm{box}},\, R(\theta_{\mathrm{yaw}}, \theta_{\mathrm{pitch}}),\, m_{\mathrm{def}},\, \mathbf{s},\, t_{\mathrm{obj}}\},
\end{equation}
where $p_{\mathrm{surf}}\in\mathbb{R}^3$ is the grounded support point,
$c_{\mathrm{box}}\in\mathbb{R}^3$ and $\mathbf{b}_{\mathrm{box}}\in\mathbb{R}_+^3$ are the ghost-box center and dimensions,
$R(\theta_{\mathrm{yaw}}, \theta_{\mathrm{pitch}})\in SO(3)$ is the inferred object orientation, constructed in a fixed world coordinate frame with roll fixed to zero under an upright-object assumption,
$m_{\mathrm{def}}$ selects the downstream fitting strategy,
$\mathbf{s}\in\mathbb{R}_+^3$ is the corresponding adaptation scale vector,
and $t_{\mathrm{obj}}$ is the semantic object description used for generation.
The deformation mode constrains $\mathbf{s}$ as rigid $\mathbf{s}=(1,1,1)$, uniform $\mathbf{s}=(\alpha,\alpha,\alpha)$, or elastic $\mathbf{s}=(s_x,s_y,s_z)$.
Thus, $G$ jointly encodes placement geometry, fitting strategy, and semantic conditioning for downstream generation and adaptation.

\subsection{Scene-Conditioned Object Prior [Stage 2]}
\label{stage2}
As shown in Figure~\ref{fig:pipeline}, given the structured fitting cues $G$, the object-prior stage produces an initial object hypothesis $O_0$. 
It first synthesizes a scene-conditioned 2D object prior and then lifts it into 3D. 
We adopt a feedforward generation-and-lifting design so that $O_0$ can be produced efficiently from scene-derived conditioning, rather than through optimization-based 3D synthesis.

\vspace{-0.7em}
\paragraph{Scene-Conditioned 2D Prior.}
Given the grounding representation $G$ and object prompt $t_{\mathrm{obj}}$ from Stage~1, we construct an initial visual hypothesis for the target object. 
Unlike prior placement methods that assume a fixed object or asset representation before placement~\cite{huang2025fireplace,feng2023layoutgpt}, ElasticFit first infers the target fitting region from the scene and uses the ghost box to define a scene-conditioned generation canvas. We employ an SDXL-style text-to-image generator~\cite{podell2023sdxl} with a scene-conditioned canvas derived from the ghost box. 
The ghost-box dimensions are defined in the box-local frame as $\mathbf{b}_{\mathrm{box}}=(b_x,b_y,b_z)$, corresponding to width, depth, and height. For front-view object generation, we set the image canvas aspect ratio according to the projected width-height proportion of the fitting region, i.e., $b_x:b_z$. This encourages the generated object to reflect the expected width-height proportions of the target region before 3D reconstruction. The object prompt $t_{\mathrm{obj}}$ specifies the desired object appearance, while the scene-conditioned canvas provides coarse structural guidance. As a result, the generated 2D object prior is biased toward the inferred fitting context rather than synthesized in isolation.

\vspace{-0.7em}
\paragraph{2D-to-3D Lifting.}
The generated 2D object image is lifted into an initial 3D object hypothesis $O_0$ using an image-to-3D model. 
In our implementation, we use TRELLIS~\cite{xiang2025structured} through a wrapper that reconstructs a 3D mesh from the generated image. 
This stage provides a scene-conditioned 3D prior that serves as a close-fit initialization. By preconditioning the object, we significantly reduce the burden of the subsequent physics-guided adaptation.

\subsection{Geometry- and Physics-Aware Fitting [Stage 3]}
\label{stage3}
While Stage~2 provides an initial prior $O_0$, the lifted mesh still requires precise alignment with the scene. Stage~3 refines the prior through a geometry- and physics-aware fitting procedure that enforces support, avoids collisions, and maximizes occupancy. Guided by the suggested deformation behavior $m_{\mathrm{def}}$, we start by employing one of the three modes described below to scale the object within the bounding box.


\begin{figure*}[t]
    \centering
    \includegraphics[width=\textwidth]{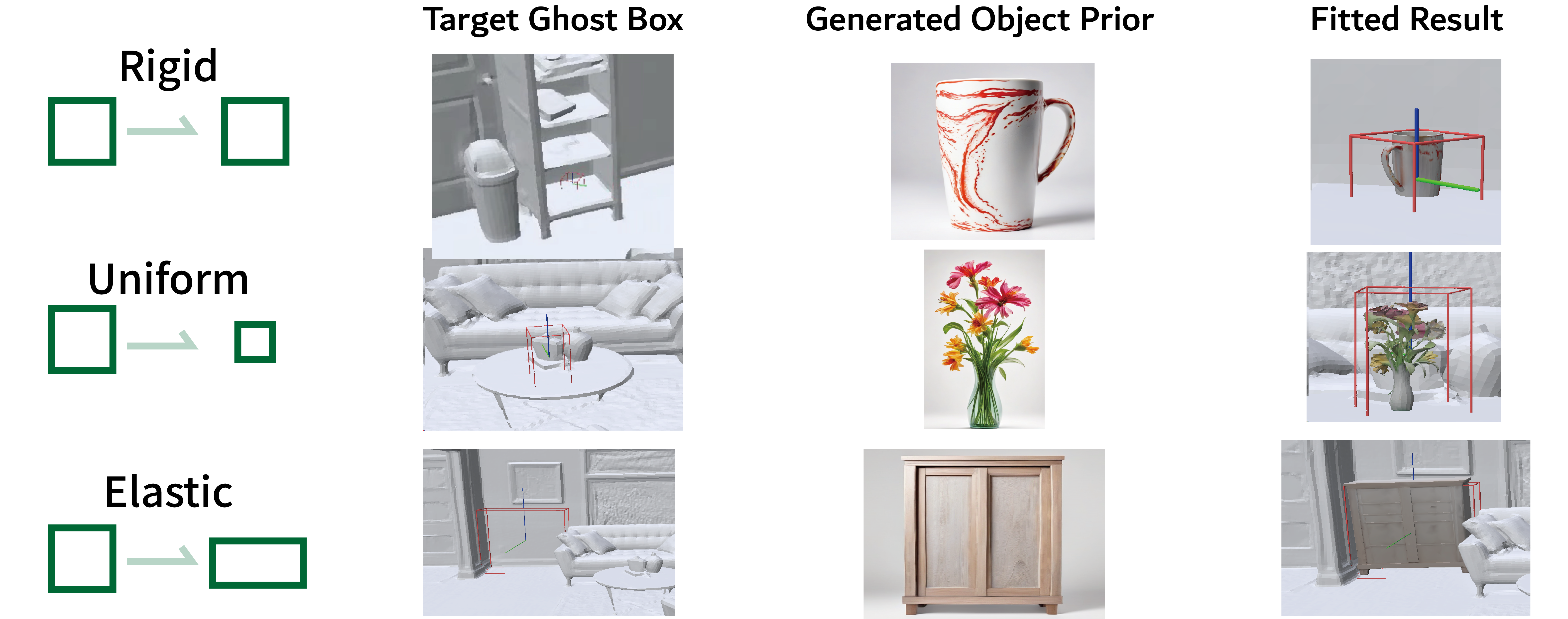}
    \caption{Visualization of the three adaptation modes in Stage~3. Each row shows the target ghost box, generated initial object, and fitted result. 
Rigid mode preserves geometry, uniform mode applies isotropic scaling, and elastic mode applies anisotropic deformation to fit the target region.}
    \label{fig:mode_example}
\end{figure*}

\vspace{-0.7em}
\paragraph{Mode-Specific Geometric Fitting.}



\textbf{Rigid} mode is used for objects whose manufactured or mechanical proportions should remain fixed. 
We preserve the object geometry and solve for a grounded, collision-aware pose using a broad-search and narrow-refine procedure: coarse proxy-box collision filtering is followed by mesh-level collision and support refinement.

\textbf{Uniform} mode is used for objects whose recognizable proportions should be preserved. 
We optimize a single global scale factor together with translation, balancing collision penalties, box constraints, support/gravity terms, and scale maximization so that the object fits the target region without changing its aspect ratio.

\textbf{Elastic} mode is used for objects that can be non-uniformly adjusted to fit constrained or gap-like regions. 
The object is anisotropically scaled along the box-local axes toward the ghost-box dimensions, while nearby scene samples are used to penalize collisions and maintain support on the target surface. 
This enables non-uniform adaptation to the available space without assuming that the entire ghost box is collision-free.

Appendix~\ref{app:fitting_details} further provides implementation details, and Algorithm~\ref{alg:stage3_fitting} summarizes the overall fitting procedure.

\vspace{-0.7em}
\paragraph{Physics-Guided Settling.}
After mode-specific fitting, we apply a lightweight rigid-body simulation in PyBullet to allow the adapted object to settle naturally in the existing scene. 
This step improves physical realism by refining the already fitted object pose under gravity and contact dynamics. 
We model the scene geometry as a static concave triangle mesh and the adapted object as a dynamic rigid body, simulating gravity, collision/contact response, friction, and damping. 
To avoid unstable contact initialization, we start the object from a slightly lifted, non-penetrating pose near the fitted placement before running the short settling simulation. 
We then perform a final refinement based on nearby upward-facing support points, applying only a conservative correction to eliminate small numerical gaps or residual contact errors. 
The final output is a fitted insertion $O^\ast$ that respects the semantic intent of $G$ while resting naturally on the supporting scene geometry.

\section{Experiments}
\label{experiments}
\subsection{Experimental Setup}

\paragraph{Dataset and Tasks.}
We evaluate ElasticFit on single-object insertion tasks, where a target object must be inserted into an existing 3D scene according to a natural-language semantic-spatial instruction while satisfying geometric and physical plausibility. 
The evaluation is designed to measure both standard placement quality and fit-aware adaptation under local scene constraints. 
We use both ReplicaCAD~\cite{szot2021habitat} and Replica~\cite{straub2019replica} for cross-dataset validation: ReplicaCAD provides clean CAD-like scenes for testing geometric soundness, while Replica provides noisy reconstructed scans for testing robustness. 
Cross-dataset results are reported in Sec.~\ref{dataset_validation}.

\vspace{-0.7em}
\paragraph{Task Construction.}
We design two task sets for evaluating single-object insertion. The \textbf{Common Placement Set} (S001--S060) contains 60 standard insertion tasks covering common spatial relations, including \texttt{near/far}, \texttt{on\_top\_of}, \texttt{left/right/front/behind}, \texttt{face\_to}, \texttt{aligned\_with}, and \texttt{against\_wall}. 
This set evaluates semantic-spatial grounding and placement quality under common relations, and is used for cross-dataset validation and baseline comparison. For experiments on this set, all methods are evaluated using the same prompts, scenes, and target object categories.


The \textbf{Fit-Critical Set} (F001--F060) contains 60 additional tasks for evaluating fit-aware insertion under challenging placement and fitting conditions. 
These tasks use the same spatial-relation vocabulary as the Common Placement Set, but add pose, contact, and local geometric adaptation requirements. 
We divide this set into two subsets: \textbf{pose-and-contact tasks}, which combine common placement relations with user-specified rotations, leaning, or lying configurations, and \textbf{gap-fitting tasks}, which require insertion into gaps, corners, wall-side spaces, or under/inside furniture.

\vspace{-0.7em}
\paragraph{Baselines.}\label{baselines}
We compare against three representative language- or vision-language-guided layout methods: \textbf{LayoutGPT}\cite{feng2023layoutgpt}, \textbf{Holodeck}\cite{yang2024holodeck}, and \textbf{LayoutVLM}\cite{sun2025layoutvlm}. We adapt them to our single-object insertion setting by keeping the input scene unchanged and allowing each method to predict or optimize only the inserted target object. In the baseline comparison, all methods receive the same scene, natural-language prompt, and fixed target object asset for each task. For clarity, we refer to these adapted implementations by their original method names in the remaining experiments.

LayoutGPT predicts the target object placement from the prompt and scene context. Holodeck uses a constraint-based search procedure to satisfy spatial relations between the target object and existing scene objects. LayoutVLM combines vision-language reasoning with differentiable constraint optimization to predict and refine the object placement. 


\label{ours-placement}
We evaluate the adapted baselines on the \textbf{Common Placement Set}, where all methods use fixed target assets for fair comparison.
For baseline comparison, \textbf{ElasticFit} is evaluated in a placement-only setting: scene-conditioned object prior generation and deformation-aware adaptation are disabled, while semantic-spatial grounding and geometry- and physics-aware refinement are retained.
This isolates placement and grounding quality from object generation and shape adaptation. The complete \textbf{ElasticFit} further includes scene-conditioned object prior generation, 3D lifting, and deformation-aware fitting, and is evaluated separately on the \textbf{Fit-Critical Set} in Sec.~\ref{fit_aware_insertion_analysis}.
All LLM/VLM-based reasoning in ElasticFit and the adapted baselines uses the same backbone, Gemini-3.1 Pro Preview.

\vspace{-0.7em}
\paragraph{Evaluation Metrics.}
We evaluate each method using four metrics. 
The first three are automatic geometry-based percentages adapted from common evaluation dimensions in 3D layout generation and object placement: spatial constraint satisfaction, collision avoidance, and physical support~\cite{huang2025fireplace,abdelreheem2025placeit3d,sun2025layoutvlm}. 
\textit{Spatial Relation Success Rate (SRS)} measures the percentage of ground-truth spatial predicates satisfied by the final insertion using checks on pose, size, orientation, and reference geometry. 
\textit{Collision-Free Rate (CFR)} measures the percentage of insertions that avoid unacceptable interpenetration, estimated by counting sampled scene points against the fitted object under a fixed tolerance.
\textit{Support Success Rate (SupR)} measures the percentage of insertions supported by the annotated surface, based on object-bottom gap and footprint support coverage. 
\textit{Contextual Plausibility Score (CPS)} follows recent model-assisted evaluation protocols~\cite{sun2025layoutvlm,huang2025fireplace}, scoring anonymized user-view screenshots and the original prompt with GPT-5.4-mini on a 1--5 scale.

\subsection{Cross-Dataset Validation}
\label{dataset_validation}


\vspace{-0.7em}
\paragraph{Protocol.}
We evaluate the robustness of ElasticFit across clean and reconstructed indoor scenes by comparing ReplicaCAD and Replica on a matched subset of the Common Placement Set. Each task uses the same prompt type, target object category, and evaluator across the two datasets. To isolate the effect of scene geometry, we use the same placement-only variant of ElasticFit as in the baseline comparison described in Sec.~\ref{baselines}.


\vspace{-0.7em}
\paragraph{Results.}
As shown in Table~\ref{tab:dataset_validation}, ElasticFit performs comparably across the two datasets, effectively bridging the gap between synthetic environments and reconstructed scans. ReplicaCAD is slightly better in support success and contextual plausibility due to cleaner CAD-like geometry, while Replica remains close despite reconstruction artifacts and inherent noise from scanned environments. This indicates robust grounding and refinement across both clean and reconstructed scenes. We conduct our experiments on Replica as it presents typical challenges of real-world AR/VR applications.





\begin{table*}[t]
\centering
\small
\setlength{\tabcolsep}{3.5pt}
\renewcommand{\arraystretch}{0.95}

\begin{minipage}[t]{0.43\textwidth}
\centering
\begin{threeparttable}
\caption{Cross-dataset validation of ElasticFit on matched placement tasks.}
\label{tab:dataset_validation}

\begin{tabular*}{\linewidth}{@{\extracolsep{\fill}}lcccc@{}}
\toprule
\textbf{Dataset} & \textbf{SRS $\uparrow$} & \textbf{CFR $\uparrow$} & \textbf{SupR $\uparrow$} & \textbf{CPS $\uparrow$} \\
\midrule
ReplicaCAD & \textbf{70.4\%} & 91.4\% & \textbf{92.7\%} & \textbf{4.58} \\
Replica    & 69.7\% & \textbf{91.7\%} & 91.7\% & 4.18 \\
\bottomrule
\end{tabular*}

\end{threeparttable}
\end{minipage}
\hfill
\begin{minipage}[t]{0.54\textwidth}
\centering
\begin{threeparttable}
\caption{Baseline comparison on the Common Placement Set.}
\label{tab:baseline_comparison}

\begin{tabular*}{\linewidth}{@{\extracolsep{\fill}}lcccc@{}}
\toprule
\textbf{Method} & \textbf{SRS $\uparrow$} & \textbf{CFR $\uparrow$} & \textbf{SupR $\uparrow$} & \textbf{CPS $\uparrow$} \\
\midrule
LayoutGPT  & 50.8\% & 73.3\% & 48.3\% & 3.72 \\
Holodeck   & 41.5\% & \textbf{95.0\%} & 41.7\% & 3.63 \\
LayoutVLM  & 49.0\% & 93.0\% & 43.9\% & 3.82 \\
\textbf{ElasticFit} & \textbf{69.7\%} & 91.7\% & \textbf{91.7\%} & \textbf{4.18} \\
\bottomrule
\end{tabular*}

\end{threeparttable}
\end{minipage}

\vspace{0.3em}
\begin{minipage}{0.95\textwidth}
\footnotesize
\textit{Note.} SRS = Spatial Relation Success Rate, CFR = Collision-Free Rate, SupR = Support Success Rate, and CPS = Contextual Plausibility Score. Higher is better for all metrics. Scores are reported in percent except CPS.
\end{minipage}

\vspace{-0.8em}
\end{table*}


\subsection{Baseline Comparison}
\label{sec:baseline_comparison}

\paragraph{Quantitative Results.}
We evaluate the adapted baselines and ElasticFit on the Common Placement Set using matched prompts, scenes, target object categories, and the same fixed target assets as the baselines.
As shown in Table~\ref{tab:baseline_comparison}, ElasticFit achieves the best spatial relation success, support success, and contextual plausibility.
It improves SRS from the strongest baseline's 50.8\% to 69.7\%, substantially increases SupR from the best baseline's 48.3\% to 91.7\%, and obtains the highest CPS of 4.18 compared with the best baseline score of 3.82.

While Holodeck and LayoutVLM obtain marginally higher CFR, this reflects a conservative strategy that places objects far from the cluttered scene geometry. Consequently, they fail to satisfy proximity and contact constraints, as evidenced by their significantly lower SRS and SupR.
ElasticFit grounds objects closer to the intended scene regions and support surfaces, improving SRS and SupR while occasionally introducing minor residual intersections on irregular geometry or tightly constrained placements.
Overall, ElasticFit maintains a high CFR while producing insertions that better satisfy spatial relations, support contact, and contextual plausibility.

\begin{figure*}[t]
    \centering
    \includegraphics[width=\linewidth]{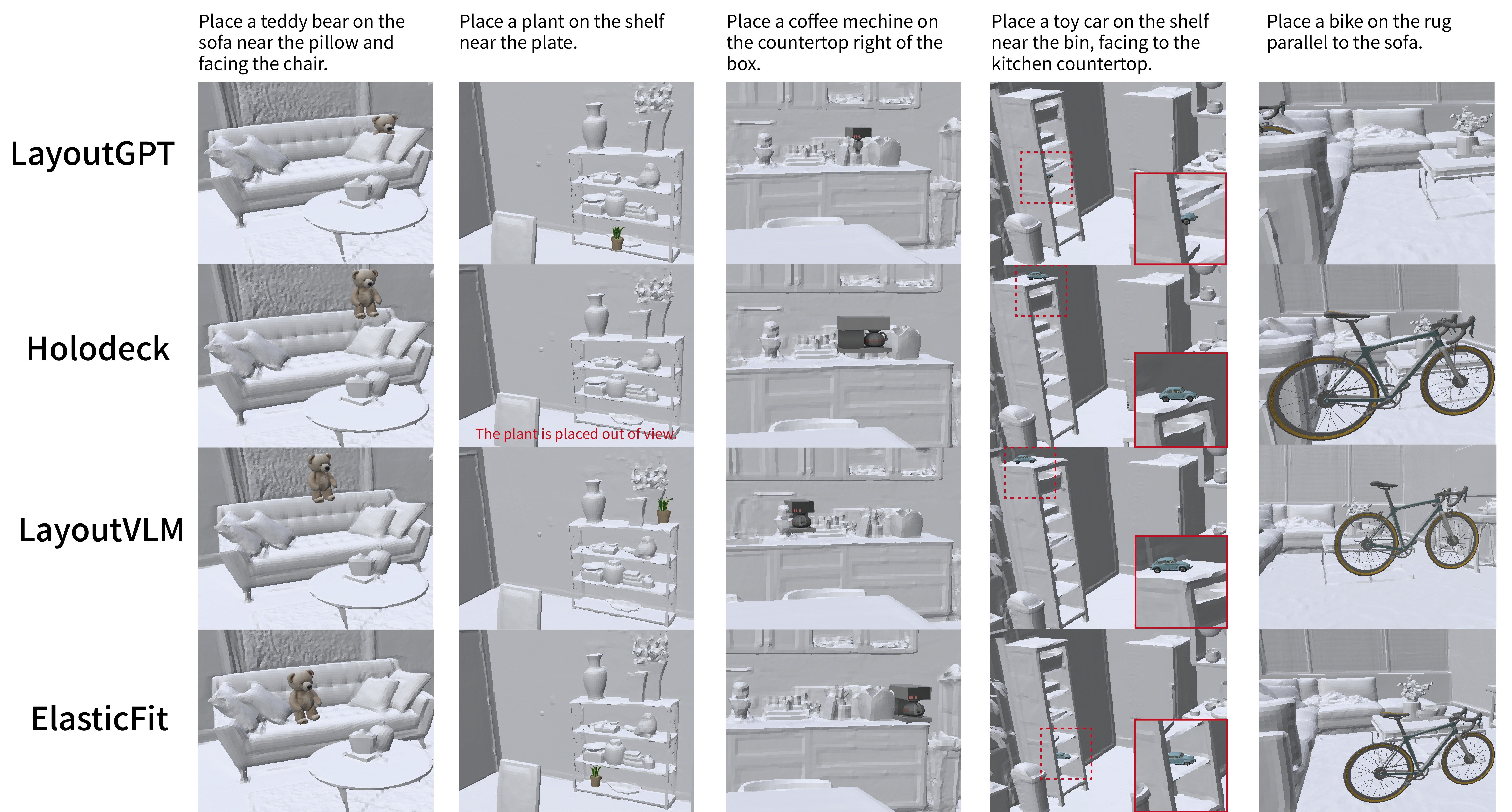}
    \caption{
    Qualitative comparison on representative Common Placement Set tasks.
    Baselines often struggle with physical refinement, ambiguous references, viewpoint-dependent relations, or irregular support geometry.
    ElasticFit produces more spatially and physically plausible placements.
    }
    \label{fig:baseline_qualitative}
\end{figure*}

\vspace{-0.7em}
\paragraph{Qualitative Results.}
Figure~\ref{fig:baseline_qualitative} shows representative examples from the Common Placement Set.
LayoutGPT often predicts plausible coarse locations, but lacks physical refinement and can produce penetrating or unsupported placements.
Holodeck improves collision avoidance, but can select incorrect references in cluttered scenes.
LayoutVLM reduces ambiguity with additional views and facing cues, but still suffers from imprecise grounding and high runtime cost; runtime comparisons are provided in Appendix~\ref{app:runtime}.
Holodeck and LayoutVLM rely on bounding-box-based refinement, which can produce floating artifacts on irregular support geometry.

A common issue across baselines is handling viewpoint-dependent spatial relations, such as \textit{in front of}, \textit{behind}, \textit{left of}, and \textit{right of}. Since users may rotate the scene during placement, these relations should follow the user's view rather than the reference object's intrinsic orientation or a fixed world frame. For example, in the coffee-machine example, a baseline may place it on the box's object-centric right side, which appears on the left from the user's current viewpoint.
ElasticFit aligns VLM reasoning with the user's current scene view and applies geometry- and physics-aware refinement, producing placements that better satisfy spatial relations, support contact, and collision avoidance.

\subsection{Fit-Aware Insertion Analysis}
\label{fit_aware_insertion_analysis}

\paragraph{Quantitative Results.}

\begin{wraptable}{r}{0.65\linewidth}
\vspace{-1.0em}
\centering
\small
\caption{Performance on the Fit-Critical Set.}
\label{tab:full_pipeline_quality}
\vspace{0.2em}
\setlength{\tabcolsep}{3pt}
\renewcommand{\arraystretch}{0.92}
\resizebox{\linewidth}{!}{
\begin{tabular}{lcccc}
\toprule
\textbf{Subset} & \textbf{SRS $\uparrow$}  & \textbf{CFR $\uparrow$} & \textbf{SupR $\uparrow$} & \textbf{CPS $\uparrow$} \\
\midrule
All tasks & 60.8\% & 71.7\% & 83.3\% & 4.57 \\
Pose-and-contact tasks & 73.3\% & 76.7\% & 80.0\% & 4.50 \\
Gap-fitting tasks & 48.3\% & 66.7\% & 86.7\% & 4.63 \\
\bottomrule
\end{tabular}
}
\vspace{-0.8em}
\end{wraptable}

Table~\ref{tab:full_pipeline_quality} summarizes the performance of the complete \textbf{ElasticFit} method on the Fit-Critical Set and its two subsets. 
Overall, ElasticFit achieves 60.8\% SRS, 71.7\% CFR, and 83.3\% SupR. 
The higher performance on pose-and-contact tasks shows that ElasticFit is effective when the main challenge is to satisfy explicit pose, orientation, support, or contact constraints. 
In this subset, ElasticFit achieves 73.3\% SRS and 76.7\% CFR, suggesting that semantic-spatial grounding and physics-guided refinement can reliably produce physically plausible insertions.

In contrast, gap-fitting tasks are more challenging, with SRS decreasing to 48.3\% and CFR to 66.7\%.
This is expected, as insertions in highly constrained spaces are sensitive to minor geometric misalignments. 
Nevertheless, SupR remains high at 86.7\%, and CPS reaches 4.63, indicating that many results are still well-supported and contextually plausible. 
This suggests that strict geometric-relation scores on constrained-space tasks should be interpreted together with contextual plausibility; we provide a complementary analysis in Appendix~\ref{app:gap_srs_failure_analysis}.

\vspace{-0.7em}
\paragraph{Qualitative Results.}
As shown in Figure~\ref{fig:teaser}, ElasticFit produces fit-aware insertions that conform to local scene geometry beyond rigid placement.

For pose-and-contact tasks, ElasticFit supports pitch-angle adjustments for \textit{leaning} and \textit{lying} configurations, enabling objects such as a teddy bear to tilt or lie naturally on a sofa. 
It also supports user-defined rotations, allowing users to specify desired object orientations more explicitly.

For gap-fitting tasks, ElasticFit adapts object geometry by elongating, compressing, or scaling generated objects to occupy narrow regions beside furniture, inside shelves, or under existing objects. 
These examples illustrate the key motivation of ElasticFit: rather than only searching for a valid pose for a fixed asset, ElasticFit adapts the inserted object to better match local spatial affordances.




\subsection{Ablation Studies}
\label{sec:ablation}

\begin{wraptable}{r}{0.57\linewidth}
\vspace{-1.5em}
\centering
\small
\caption{Ablation study on gap-fitting tasks.}
\label{tab:ablation_adaptive}
\setlength{\tabcolsep}{3pt}
\renewcommand{\arraystretch}{0.92}
\resizebox{\linewidth}{!}{
\begin{tabular}{lccccc}
\toprule
\textbf{Setting} 
& \textbf{SRS $\uparrow$}  
& \textbf{CFR $\uparrow$} 
& \textbf{SupR $\uparrow$}
& \textbf{CPS $\uparrow$}
& \textbf{Pen. $\downarrow$} \\
\midrule
ElasticFit 
& \textbf{47.8\%} 
& \textbf{76.7\%} 
& \textbf{83.3\%}  
& \textbf{4.63} 
& \textbf{1.08\%} \\

w/o physics settle 
& 46.7\% 
& \textbf{76.7\%} 
& 80.0\%  
& 4.60
& 1.45\% \\

w/o geom.-phys. fitting
& \textbf{47.8\%} 
& 36.7\%  
& \textbf{83.3\%} 
& 4.23
& 6.09\% \\
\bottomrule
\end{tabular}
}
\end{wraptable}

Table~\ref{tab:ablation_adaptive} evaluates the contribution of geometry- and physics-aware fitting on gap-fitting tasks. It uses a separate controlled run with a different input view from Table~\ref{tab:full_pipeline_quality}, while sharing the same view across all variants. The second row removes only final settling, while the third removes mode-specific fitting and uses the initial ghost-box fit.
We also report penetrating vertex ratio, the fraction of object vertices estimated to penetrate the scene; lower is better.
Removing only the final physics-based settling leads to a decrease in SupR from 83.3\% to 80.0\%, and an increase in penetrating vertex ratio from 1.08\% to 1.45\%. This confirms that while geometric fitting identifies the correct region, the dynamic simulation is necessary to resolve small-scale issues.

In contrast, removing geometry- and physics-aware fitting in Stage~3 substantially degrades geometric feasibility. 
Although SRS remains similar and SupR is unchanged, CFR drops from 76.7\% to 36.7\%, and penetrating vertex ratio increases from 1.08\% to 6.09\%. 
This indicates that the initial ghost-box fit can often identify a semantically reasonable target region and supported placement, but is insufficient for collision-free insertion in constrained spaces. 
The lower CPS score further suggests that severe geometric misfit can also reduce the perceived plausibility of the final result.





\section{Conclusion and Discussion}
ElasticFit formulates 3D object insertion as fit-aware adaptation, grounding generated objects in pre-existing scenes while adapting them to local geometric constraints. 
Rather than directly predicting rigid placements, ElasticFit infers scene-grounded fitting cues that connect semantic intent, object generation, and deformation-aware fitting. 
By combining view-aligned VLM grounding, scene-conditioned object priors, and geometry- and physics-aware fitting, ElasticFit improves spatial relation and support success over language- and vision-language-guided baselines, while enabling constrained insertions beyond placement-only methods. 
These results highlight generative adaptation as a bridge between high-level semantic instructions and physically grounded 3D scene realization.

Despite these gains, our current elastic mode relies on anisotropic scaling. While effective for many categories of objects, true non-rigid deformation (e.g., cloth draping over a chair or pillow compressed into a corner) remains an open challenge. Moreover, because of proximity constraints, sometimes the support constraint is satisfied, but objects are left partially hanging over a surface edge. Incorporating explicit containment constraints could ensure that objects are fully within the supporting surface.
Future work could also explore multi-object insertion, where reasoning should account for multiple simultaneous object placements.

\bibliographystyle{unsrtnat}
\bibliography{references}


\clearpage
\appendix

\section*{Appendix Overview}

This appendix provides additional details, analyses, and qualitative results for ElasticFit.

In Appendix~\ref{app:system_limitations}, we discuss system limitations and representative failure cases, including directional errors, deformation artifacts, and concave support ambiguities.

In Appendix~\ref{app:comparison}, we provide a design-level comparison with closely related object placement and layout methods, with a particular focus on FirePlace as the closest prior work to our setting.

In Appendix~\ref{app:implementation}, we provide implementation details for the three stages of ElasticFit: semantic-spatial grounding, scene-conditioned object prior generation, and geometry- and physics-aware fitting. We also include the structured VLM output schema, prompt organization, fitting objective, and the overall Stage~3 fitting algorithm.

In Appendix~\ref{app:tasks}, we describe the task sets and prompt construction used in our evaluation, including the Common Placement Set and the Fit-Critical Set.

In Appendix~\ref{app:metrics}, we define the evaluation metrics used in the main paper, including Spatial Relation Success Rate, Collision-Free Rate, Support Success Rate, and Contextual Plausibility Score.

In Appendix~\ref{app:baselines}, we describe how the baseline methods are adapted to the single-object insertion setting and provide additional runtime and implementation details for comparison.

In Appendix~\ref{app:additional_qualitative}, we show additional qualitative examples for baseline comparisons and fit-aware insertion results.

In Appendix~\ref{app:asset_licenses}, we document the third-party 3D assets used in our experiments and their licenses.

\section{System Limitations}
\label{app:system_limitations}
\paragraph{Evaluation Scope.}
Our task suite was developed alongside the method, so the reported
results should be interpreted within this controlled setting.
We do not quantify repeated-run variability.
Our experiments use dataset-provided semantic instance annotations,
and robustness to errors in predicted annotations remains untested.

\subsection{Complementary Interpretation of SRS and CPS on Gap-Fitting Tasks}
\label{app:gap_srs_failure_analysis}

The gap-fitting subset contains 30 tasks and has a lower SRS of 48.3\%. 
This result should be interpreted together with CPS, since the two metrics measure different aspects of insertion quality. 
SRS evaluates whether specified geometric predicates are strictly satisfied, while CPS evaluates the overall contextual plausibility of the rendered result. 
Among the 25 gap-fitting cases with $\mathrm{SRS}<1$, 24 cases still receive a high CPS score ($\mathrm{CPS}\geq 4$). 
This suggests that many SRS-unsatisfied cases correspond to strict predicate-level deviations, especially orientation-sensitive checks in constrained spaces, rather than clearly implausible insertions.

\begin{table}[h]
\centering
\caption{
Diagnostic breakdown of SRS-unsatisfied cases on the gap-fitting subset.
The gap-fitting subset contains 30 tasks; this table analyzes the 25 cases with $\mathrm{SRS}<1$.
Categories are diagnostic and not mutually exclusive.
}
\label{tab:gap_failure_breakdown}
\small
\setlength{\tabcolsep}{4pt}
\renewcommand{\arraystretch}{1.15}
\begin{tabular}{@{}p{0.40\linewidth}c p{0.46\linewidth}@{}}
\toprule
\textbf{Diagnostic category} & \textbf{Count / 25} & \textbf{Explanation} \\
\midrule
SRS not satisfied, but visually plausible 
($\mathrm{CPS}\geq 4$)
& 24 (96.0\%) 
& These cases do not fully satisfy strict geometric predicates, but their rendered placements remain contextually plausible according to CPS. \\

\texttt{parallel\_to} check not satisfied
& 19 (76.0\%)
& These cases contain an unsatisfied \texttt{parallel\_to} relation, indicating that orientation constraints are a common source of strict SRS misses in constrained regions. \\
\bottomrule
\end{tabular}
\end{table}


\subsection{Representative Failure Cases}
\begin{figure}[h]
    \centering
    \includegraphics[width=\linewidth]{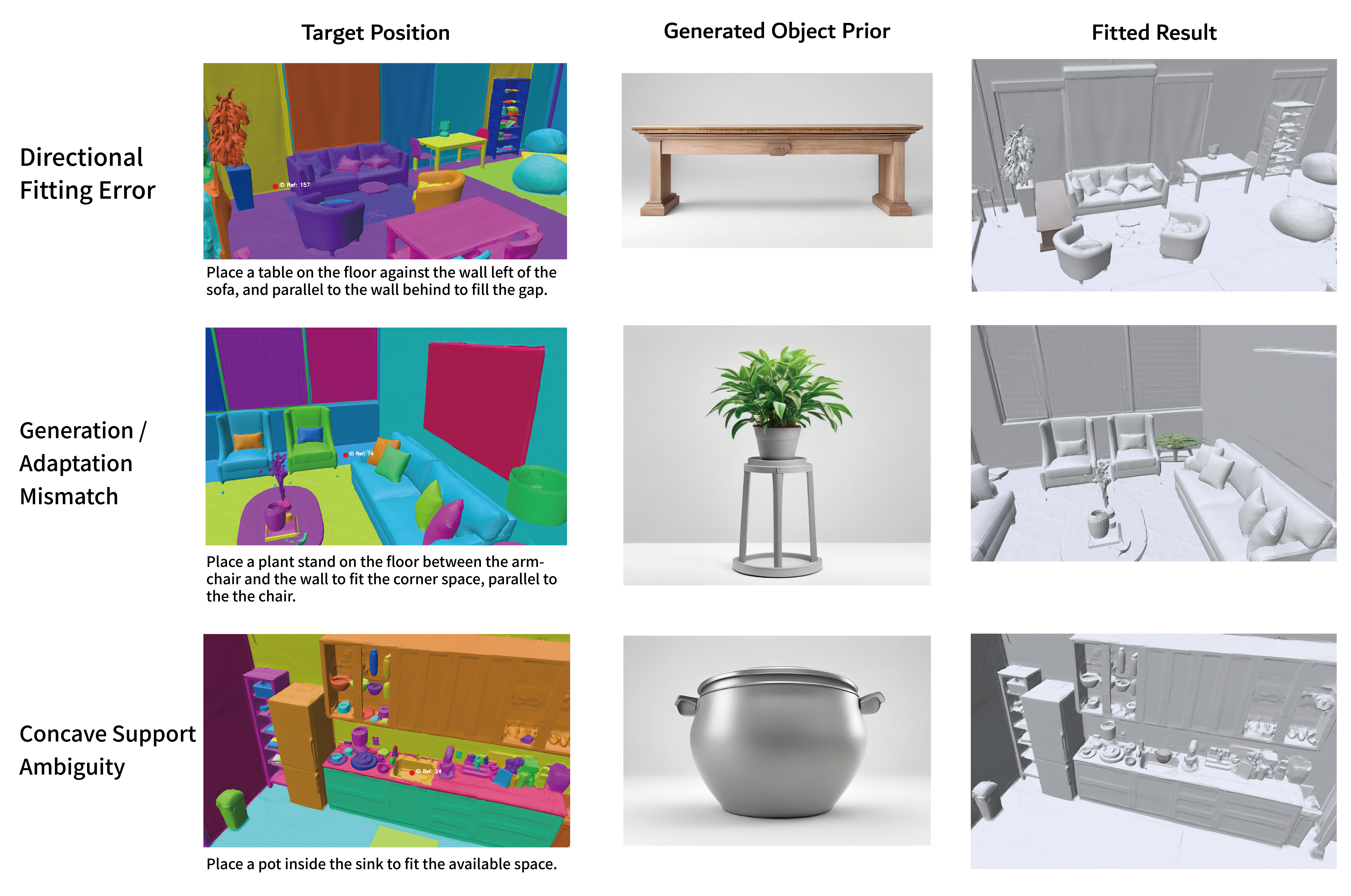}
    \caption{
    Representative failure cases of ElasticFit.
    Top: directional fitting errors arise from ambiguous reference geometry or fill directions.
    Middle: generation--adaptation mismatch occurs when the object prior and subsequent deformation do not match the target fitting region.
    Bottom: concave support ambiguity occurs in sink-like geometry, where rim or surface support can be mistaken for the intended interior placement.
    }
    \label{fig:failure_cases}
\end{figure}

Figure~\ref{fig:failure_cases} shows representative failure cases of ElasticFit, highlighting limitations in directional fitting, deformation-aware adaptation, and concave support reasoning.

\begin{table}[h]
\centering
\caption{
Failure attribution on the Fit-Critical Set.
Each failed task is assigned to its dominant failure source.
Percentages are computed over all 60 Fit-Critical tasks.
}
\label{tab:failure_attribution}
\small
\setlength{\tabcolsep}{4pt}
\renewcommand{\arraystretch}{1.15}
\begin{tabular}{@{}p{0.22\linewidth}c p{0.5\linewidth}@{}}
\toprule
\textbf{Failure source} & \textbf{Count / 60} & \textbf{Description} \\
\midrule
Directional fitting error 
& \makecell[l]{2 (3.3\%)} 
& Incorrect yaw, pitch, alignment, or gap-fill direction. \\

Generation / adaptation mismatch 
& \makecell[l]{6 (10.0\%)} 
& Object prior or deformation mismatches the target fitting region. \\

\makecell[l]{Concave support\\ambiguity}
& \makecell[l]{1 (1.7\%)} 
& Sink-like concave geometry can confuse inside placement with rim or surface support. \\
\bottomrule
\end{tabular}
\end{table}

Table~\ref{tab:failure_attribution} summarizes the dominant visible failure sources observed in the Fit-Critical Set. We discuss representative cases below.

First, directional fitting errors can occur even when the target support point is semantically corrected. 
ElasticFit verifies whether the predicted target point lies on the intended support object using the semantic ID buffer and, when necessary, corrects it to a valid point on the support object. 
However, this correction only ensures that the point belongs to the intended semantic object; it does not guarantee that the point lies on the intended support surface, such as the top of a shelf rather than a nearby side surface. 
This makes the fitting direction sensitive to viewpoint and surface visibility, and may lead to floating or offset placements when the visible image evidence is ambiguous. 
Moreover, even when the support point is corrected back to the intended support surface, the correction does not fully resolve the orientation or fill direction of the target ghost box. 
For example, in the first row, the instruction asks for a table to fill the gap along the wall. 
Although the support point is corrected back to the floor, the current orientation heuristic estimates parallel or perpendicular alignment from the long and short axes of the reference object. 
When the reference object has similar horizontal extents or ambiguous dominant axes, the inferred fill direction can be incorrect. 
As a result, the fitted object is placed in a semantically related region but fails to occupy the intended wall-side gap.

Second, generation--adaptation mismatch may arise from the combination of soft object-prior control and elastic fitting.
ElasticFit uses the target ghost-box canvas and coarse topology cues, such as \textit{wide horizontal} or \textit{tall vertical}, to encourage the generated object prior to match the intended aspect ratio.
However, these cues do not impose hard geometric constraints on the generative model, and the generated prior may still underfill the target canvas or follow shape biases from the model's training distribution.
When this prior is subsequently adapted through axis-dependent scaling in the local frame of the target ghost box, elastic fitting can amplify the mismatch, producing visually implausible shapes for objects with strict semantic or structural proportions.
As shown in the second row, the generated plant prior does not fully match the intended elongated fitting region, and the fitted result exhibits an unnatural deformation.
This suggests the need for stronger spatial conditioning, such as ControlNet-style or mask/box-conditioned generation, together with a deformation-tolerance prior that determines which object categories can safely undergo elastic adaptation.

Third, concave support ambiguity can occur in sink-like geometry, where the intended placement is \textit{inside} a basin rather than \textit{on} a surface. 
In the sink example in the third row, support and collision refinement can treat the rim or nearby upward-facing geometry as the effective support instead of the basin interior. 
This may lift the object away from the intended interior region, causing the final placement to appear floating or insufficiently contained.

These cases suggest several directions for future work, including stronger spatial conditioning for object-prior generation, more robust orientation reasoning for ambiguous reference geometry, category-aware deformation constraints, and iterative verification of the fitted result against the original instruction and local scene geometry.


\subsection{Broader Impacts}
ElasticFit may support AR/VR authoring, interior design, simulation, and embodied-AI scene construction by reducing manual effort for placing and adapting objects in existing 3D scenes. Potential risks include generating misleading or unrealistic scene edits, producing physically unsafe suggestions if used in real-world design workflows without verification, and inheriting biases or artifacts from external VLMs and generative models. The system is intended for research use and should not be used for safety-critical physical planning without additional verification.

\section{Design-Level Comparison with Closely Related Work}
\label{app:comparison}
This section provides a design-level comparison between ElasticFit and closely related object placement methods, with a particular focus on FirePlace~\cite{huang2025fireplace}, which is the closest prior work to our setting. FirePlace grounds multimodal language-model reasoning in fine-grained scene geometry and refines the placement of a given 3D object using geometric constraints. However, it assumes that the inserted object is a pre-existing fixed asset. In contrast, ElasticFit studies a complementary fit-aware insertion setting, where the system must reason not only about where the object should be placed, but also about how the object should be generated or adapted to satisfy local scene constraints.

\begin{table}[t]
\centering
\caption{Design-level comparison between FirePlace and ElasticFit.}
\label{tab:fireplace_comparison}
\small
\setlength{\tabcolsep}{5pt}
\renewcommand{\arraystretch}{1.18}
\begin{tabularx}{\linewidth}{@{}p{0.20\linewidth}XX@{}}
\toprule
\textbf{Aspect} & \textbf{FirePlace} & \textbf{ElasticFit} \\
\midrule
Target object
& Pre-existing fixed asset
& Generated or fixed object \\
\midrule
Main objective
& Place a given object
& Generative ``make-it-fit'' insertion \\
\midrule
Adaptation
& Pose-level placement
& Rigid, uniform, or elastic fitting \\
\midrule
Constrained spaces
& Limited by fixed-asset geometry
& Designed for gaps, corners, wall-side, under/inside spaces \\
\midrule
Grounding
& Fine-grained geometric placement constraints
& Grounded fitting representation \(G\) with support, ghost box, orientation, mode, scale, and prompt \\
\midrule
Physical refinement
& Geometry-aware collision/support refinement
& Geometry-aware fitting plus physics-guided settling \\
\bottomrule
\end{tabularx}
\end{table}

Table~\ref{tab:fireplace_comparison} summarizes the key design differences. 
FirePlace is closer to ElasticFit than layout-level methods such as LayoutGPT, Holodeck, and LayoutVLM because it grounds multimodal reasoning in fine-grained scene geometry and refines object placement using geometric constraints. 
However, FirePlace assumes a pre-existing fixed asset and optimizes its placement, while ElasticFit targets a generative ``make-it-fit'' setting in which the object may also need to be generated or adapted to satisfy local scene constraints. 
Since our Fit-Critical Set is specifically designed for cases where fixed assets may not satisfy the available space, a direct numerical comparison on this set would evaluate FirePlace outside its intended setting. 
We therefore treat FirePlace as a design reference rather than a direct full-pipeline baseline.

ElasticFit extends this line of work by introducing a scene-grounded fitting representation that connects VLM reasoning to both generative object creation and downstream geometric fitting. 
Instead of first selecting a fixed mesh and optimizing only its pose, ElasticFit grounds the intended insertion region through a support point, ghost box, orientation, deformation mode, and object prompt. 
This representation allows the object prior to be generated or selected according to the target scene context, and then fitted using rigid placement, uniform scaling, or elastic adaptation. 
This design enables insertion in cases where the object must fit into a narrow gap, occupy a wall-side region, or adapt to a constrained space under or inside existing furniture.

\paragraph{Task compatibility with FirePlace.}
Although the Fit-Critical Set is beyond FirePlace's fixed-asset setting, parts of the Common Placement Set overlap with its intended scope. 
In particular, the distance and alignment subsets, comprising 24 of the 60 Common Placement tasks, involve standard fixed-asset placement on support surfaces and do not require object generation or shape adaptation. 
These tasks represent the closest overlap between FirePlace and ElasticFit.

Under our placement-only variant, ElasticFit disables scene-conditioned object generation and deformation-aware adaptation, while retaining view-aligned VLM grounding and geometry- and physics-aware refinement. 
This setting serves as a proxy for evaluating placement quality in the fixed-asset paradigm. 
On the full Common Placement Set, this variant achieves 69.7\% SRS and 91.7\% SupR, indicating strong geometric grounding and support accuracy in a fixed-asset comparison setting.

The remaining Common Placement subsets involve viewpoint-dependent spatial relations, explicit orientation reasoning, or contact-heavy placement near surrounding geometry. 
We therefore use the Common Placement Set for baseline-compatible quantitative comparison, and the Fit-Critical Set to evaluate ElasticFit's additional generative and adaptive capabilities.

\begin{table}[t]
\centering
\caption{Design-level comparison of object insertion and layout methods.}
\label{tab:method_comparison}
\fontsize{8.7}{10}\selectfont
\setlength{\tabcolsep}{2.5pt}
\renewcommand{\arraystretch}{1.2}
\begin{tabular}{@{}lcccc@{}}
\toprule 
\textbf{\makecell[l]{Method}} 
& \textbf{\shortstack{Target\\Object}} 
& \textbf{\shortstack{Placement\\Strategy}} 
& \textbf{\shortstack{\makecell[l]{Adaptation}}} 
& \textbf{\shortstack{Constrained-Space\\Insertion}} \\
\midrule
LayoutGPT   
& Fixed / retrieved 
& Layout pose direct prediction
& Rigid only
& No \\
Holodeck    
& Fixed / retrieved 
& Layout constraints / box solver 
& Rigid only 
& No \\
LayoutVLM   
& Fixed / retrieved 
& Visual layout / box solver
& Rigid only 
& No \\
FirePlace   
& Fixed asset 
& Fine-grained geom. constraints 
& Rigid only
& Fixed-asset only \\
\midrule
\textbf{\makecell[l]{ElasticFit}} 
& \textbf{\makecell[c]{Generated / fixed}} 
& \textbf{\makecell[c]{Grounded rep. \(G\)\\+ geom.-phys. fitting}} 
& \textbf{\makecell[c]{Rigid / Uniform / Elastic}} 
& \textbf{\makecell[c]{Yes}} \\
\bottomrule
\end{tabular}
\end{table}


\paragraph{Comparison with representative methods.}
For broader context, Table~\ref{tab:method_comparison} summarizes the design-level differences between ElasticFit and representative object insertion or layout methods. 
The comparison focuses on whether the target object is fixed, retrieved, or generated; whether placement is predicted directly or optimized through geometric fitting; whether the method adapts object geometry; and whether it is designed for constrained-space insertion where a fixed asset may not satisfy the available space.

LayoutGPT, Holodeck, and LayoutVLM primarily operate in fixed-asset or retrieved-asset layout settings. 
They can reason about object placement at the scene or bounding-box level, but they do not modify the target object's geometry to satisfy local spatial constraints. 
FirePlace is closer to ElasticFit because it uses fine-grained geometric reasoning to place a given object in an existing scene, but it still assumes a pre-existing fixed asset and does not support object generation or elastic adaptation. 
ElasticFit extends this direction by first grounding a target fitting region and then generating or adapting the object through rigid, uniform, or elastic fitting.

\section{Implementation Details}
\label{app:implementation}
This appendix details the implementation of ElasticFit's three stages: VLM grounding, scene-conditioned object prior generation, and mode-specific fitting and physics-based settling.

\subsection{Stage 1: Semantic-Spatial Grounding}
\label{app:stage1_details}

We provide additional details for the semantic-spatial grounding stage described in Sec.~\ref{stage1}, focusing on the structured VLM output, semantic correction, ghost-box construction, orientation resolution, and prompt organization.

\paragraph{VLM Output Schema.}
Stage~1 queries the VLM with the clean rendered scene image, the Set-of-Mark image, the visible object list, and the user instruction. 
The VLM is constrained to return a single JSON object following the schema in Table~\ref{tab:stage1_json_schema}. 
This schema separates spatial grounding, orientation reasoning, adaptation-mode selection, contact handling, gap filling, and object-generation conditioning.

\begin{table}[h]
\centering
\caption{Structured JSON output predicted by Stage~1.}
\label{tab:stage1_json_schema}
\resizebox{\linewidth}{!}{
\begin{tabular}{lll}
\toprule
\textbf{Field} & \textbf{Type} & \textbf{Description} \\
\midrule
\texttt{anchor\_object\_id} & int & Main reference object for spatial reasoning. \\
\texttt{support\_object\_id} & int & Object expected to support the inserted object. \\
\texttt{target\_uv} & $[u,v]$ & Image-space target point for the object's bottom center. \\
\texttt{look\_at\_uv} & $[u,v]$ & Optional point that the object front should face. \\
\texttt{top\_uv} & $[u,v]$ & Optional top contact point for leaning or tilted poses. \\
\texttt{scale} & $[w,d,h]$ & Object size in the default upright pose. \\
\texttt{deformation\_tag} & int & Adaptation mode: rigid, elastic, or uniform. \\
\texttt{pose\_tag} & int & Pose mode: upright, lying flat, or upside down. \\
\texttt{orientation\_relation\_tag} & int & Default, parallel/aligned, or perpendicular relation. \\
\texttt{orientation\_reference\_object\_id} & int & Reference object for orientation alignment. \\
\texttt{yaw\_offset\_degrees} & float & Explicit user-specified yaw offset. \\
\texttt{pitch\_offset\_degrees} & float & Explicit user-specified pitch offset. \\
\texttt{is\_snapped} & bool & Whether the object should be flush against another object. \\
\texttt{snap\_object\_id} & int & Object used for horizontal snapping. \\
\texttt{enable\_gap\_filling} & bool & Whether the instruction requests filling available space. \\
\texttt{fill\_front/back/left/right} & bool & Local directions for gap expansion. \\
\texttt{place\_object\_prompt} & string & Prompt for isolated object generation. \\
\texttt{place\_object\_name} & string & Short object name for logging and visualization. \\
\bottomrule
\end{tabular}
}
\end{table}

\paragraph{Semantic Correction.}
The VLM-predicted target point may fall slightly outside the intended support surface. 
Let $\hat{q}=(\hat{u},\hat{v})$ be the predicted target pixel and $\hat{s}$ the predicted support object ID. 
If the ID buffer does not match the expected support object, we replace $\hat{q}$ with the nearest visible pixel on the intended support object:
\begin{equation}
    q^\star =
    \arg\min_{q \in \Omega(\hat{q})}
    \| q - \hat{q} \|_2
    \quad
    \mathrm{s.t.}
    \quad
    B_{\mathrm{id}}(q) = \hat{s}.
\end{equation}
The grounded support point is then obtained from the XYZ buffer:
\begin{equation}
    p_{\mathrm{surf}} = B_{\mathrm{xyz}}(q^\star).
\end{equation}

\paragraph{Ghost-Box Construction.}
The predicted object scale is interpreted in the default upright pose, $b_{\mathrm{box}}=(w,d,h)$. 
Given the grounded support point $p_{\mathrm{surf}}$ and orientation $R(\theta_{\mathrm{yaw}},\theta_{\mathrm{pitch}})$, we place the ghost box by anchoring the center of its bottom face at $p_{\mathrm{surf}}$:
\begin{equation}
    c_{\mathrm{box}}
    =
    p_{\mathrm{surf}}
    -
    R(\theta_{\mathrm{yaw}}, \theta_{\mathrm{pitch}})
    \begin{bmatrix}
    0 \\ 0 \\ -h/2
    \end{bmatrix}.
\end{equation}
This convention keeps the object grounded when height or pitch changes.

\paragraph{Orientation Resolution.}
ElasticFit resolves object orientation using a priority order. 
Explicit yaw instructions in the user prompt override inferred facing cues. 
Otherwise, yaw is estimated from \texttt{look\_at\_uv} when available, or from the oriented bounding box of a reference object for relations such as ``parallel to'', ``aligned with'', or ``perpendicular to''. 
For leaning configurations, pitch is estimated from the vector between the support point and the predicted \texttt{top\_uv}; for lying configurations, the base pitch is set to $-\pi/2$.

\paragraph{Prompt Structure.}
Table~\ref{tab:stage1_prompt_structure} summarizes the Stage~1 prompt organization. 
The prompt specifies the input view, coordinate convention, visible object list, placement rules, adaptation-mode definitions, snapping and gap-filling logic, and the required JSON output schema. 
The full prompt template will be released with the code for reproducibility.


\begin{table}[h]
\centering
\caption{Prompt structure for Stage~1 VLM reasoning.}
\label{tab:stage1_prompt_structure}
\resizebox{\linewidth}{!}{
\begin{tabular}{p{0.34\linewidth}p{0.59\linewidth}}
\toprule
\textbf{Prompt block} & \textbf{Purpose} \\
\midrule
Role and task &
Defines the VLM as a 3D spatial reasoning stage for placing a new object in an existing 3D scene. \\

Scene context and inputs &
Provides image resolution, normalized image coordinates, visible object IDs, object metadata, and the user instruction. \\

Reference and scale rules &
Selects a valid anchor object and predicts the target object dimensions in its default upright pose. \\

Coordinates and placement rules &
Predicts the target support point, optional facing point, and optional top-contact point for leaning or lying poses. \\

Physical attributes and orientation &
Predicts adaptation mode, pose tag, orientation relation, yaw offset, pitch offset, and orientation reference object. \\

Snapping and gap filling &
Handles contact-heavy instructions such as \textit{against} or \textit{flush with}, and fill-space instructions such as \textit{fill}, \textit{occupy}, or \textit{fit}. \\

Image-generation prompt &
Produces an isolated-object prompt and a short object name for scene-conditioned generation. \\

Output schema &
Enforces a strict JSON object with predefined keys and no additional text. \\
\bottomrule
\end{tabular}
}
\end{table}

\subsection{Stage 2: Scene-Conditioned Object Prior}
\label{app:stage2_details}

This section provides additional implementation details for Stage~2 beyond the high-level description in Sec.~\ref{stage2}. 
We focus on prompt construction, topology-aware canvas sizing, and generation controls used to produce stable object-centric images for 3D lifting.

\paragraph{Prompt Construction.}
The object prompt predicted by Stage~1 is converted into an object-centric image-generation prompt by appending a canonical description:
\begin{quote}
\small
\texttt{[object\_description], front view, full object visible, centered, upright, isolated single object, clean light gray studio background, realistic product photo, filling the entire frame from edge to edge, perfectly fitted, extremely large relative scale within canvas, vivid colors, high contrast against the background}
\end{quote}
These attributes reduce background clutter, cropping, and multi-object generations before image-to-3D lifting.

\paragraph{Topology-Aware Canvas and Prompt Tags.}
Given the ghost-box dimensions $\mathbf{b}_{\mathrm{box}}=(b_x,b_y,b_z)$ predicted by Stage~1, we set the front-view canvas aspect ratio using the width-height proportion of the fitting region:
\begin{equation}
    r = \frac{b_x}{b_z}.
\end{equation}
Starting from a target area of $768^2$ pixels, we compute a canvas size whose aspect ratio matches $r$, then round the width and height to the nearest multiple of 64 and clamp both dimensions to the supported range $[512,1024]$. 
For elastic objects, we also prepend a coarse topology tag to the prompt: \texttt{wide horizontal} if $r>1.2$, and \texttt{tall vertical} if $r<0.8$. 
No topology tag is added for rigid or uniform modes.

\paragraph{Negative Prompt.}
We use a negative prompt to suppress multiple-object layouts, cropping, indoor backgrounds, and low-quality artifacts, with terms such as:
\begin{quote}
\small
\texttt{multiple objects, cropped, partial object, scene, room, floor, wall, table, sofa, bed, shelf, background clutter, blurry, deformed, duplicate, text, watermark.}
\end{quote}
This helps keep the generated image object-centric for subsequent 3D reconstruction.



\subsection{Stage 3: Geometry- and Physics-Aware Fitting}
\label{app:fitting_details}

This section provides implementation details for the fitting and physics-guided settling stage described in Sec.~\ref{stage3}. 
All fitting procedures operate in the ghost-box local frame defined by the predicted center, dimensions, yaw, and pitch from Stage~1. 
The scene is represented by dense surface samples with normals, cropped around the ghost box to reduce computation while preserving local support and collision geometry.

\begin{table}[h]
\centering
\caption{Summary of the three adaptation modes in Stage~3.}
\label{tab:adaptation_modes}
\begin{tabular}{llll}
\toprule
\textbf{Mode} & \textbf{Scale} & \textbf{Main variable} & \textbf{Solver} \\
\midrule
Rigid & $(1,1,1)$ & Translation near ghost box & Heuristic search \\
Uniform & $(\alpha,\alpha,\alpha)$ & Translation and global scale & Gradient optimization \\
Elastic & $(s_x,s_y,s_z)$ & Box-local extents & Box-aligned expansion \\
\bottomrule
\end{tabular}
\end{table}

\paragraph{Mode-Specific Fitting.}
Rigid fitting uses a broad-to-narrow search: candidate anchors are first filtered by proxy-box collision and support checks, then shortlisted candidates are evaluated using mesh-level penetration statistics. 
Uniform fitting optimizes translation and a single isotropic scale factor while retaining the best safe checkpoint. 
Elastic fitting searches for a maximum empty box in the ghost-box local frame by expanding dense XY seeds along local axes under collision and support checks, then maps the generated mesh into the selected box. Algorithm~\ref{alg:stage3_fitting} summarizes the overall fitting procedure before the detailed objective and checks are described below.







\begin{algorithm}[t]
\caption{Geometry- and Physics-Aware Fitting}
\label{alg:stage3_fitting}

\KwIn{Initial mesh $O_0$, grounding representation $G$, scene samples $P$}
\KwOut{Final inserted object $O^\ast$}

Transform $O_0$ and local scene samples into the ghost-box frame defined by $G$\;
Crop nearby support and collision geometry from $P$\;

\uIf{$m_{\mathrm{def}} = \mathrm{rigid}$}{
    Search for a grounded pose while preserving object scale\;
}
\uElseIf{$m_{\mathrm{def}} = \mathrm{uniform}$}{
    Optimize translation and isotropic scale under geometric and support terms\;
}
\uElseIf{$m_{\mathrm{def}} = \mathrm{elastic}$}{
    Find a collision-aware box-local region and apply axis-dependent scaling\;
}

Select the best fitted candidate using penetration and support statistics\;
Run a short rigid-body settling simulation from a lifted initialization\;
Apply a conservative support-aware grounding correction\;

\Return{$O^\ast$}
\end{algorithm}

\paragraph{Optimization Objective.}
For optimization-based fitting, the objective combines collision, box-bound, gravity, centering, and scale terms:
\begin{equation}
    \mathcal{L}
    =
    \lambda_{\mathrm{col}}\mathcal{L}_{\mathrm{col}}
    +
    \lambda_{\mathrm{box}}\mathcal{L}_{\mathrm{box}}
    +
    \lambda_{\mathrm{grav}}\mathcal{L}_{\mathrm{grav}}
    +
    \lambda_{\mathrm{center}}\mathcal{L}_{\mathrm{center}}
    -
    \lambda_{\mathrm{scale}}\alpha .
\end{equation}
Here, $\mathcal{L}_{\mathrm{col}}$ penalizes point-to-plane penetration against nearby scene samples, $\mathcal{L}_{\mathrm{box}}$ penalizes violations of the ghost-box bounds, $\mathcal{L}_{\mathrm{grav}}$ encourages lower physically grounded placements, and $\mathcal{L}_{\mathrm{center}}$ keeps the object near the predicted target region. 
For uniform mode, $\alpha$ is the optimized isotropic scale factor. 
Rigid mode uses heuristic search rather than this objective, while elastic mode uses box-local expansion instead of gradient optimization.

\paragraph{Collision and Support Checks.}
Collision is estimated using both coarse proxy-box tests and mesh-level point-to-plane checks. 
For mesh-level checks, each object vertex queries nearby scene samples and computes a local signed distance using scene normals. 
Penetration statistics such as maximum penetration, mean penetration, and penetration ratio are used to select or reject candidates. 
Support reasoning uses upward-facing scene samples under the object footprint and robust height estimates from top-ranked support points, reducing sensitivity to noisy scan geometry.

\paragraph{Physics-Guided Settling and Final Support Clamp.}
After mode-specific fitting, ElasticFit applies a short PyBullet rigid-body settling step. 
The scene is loaded as a static concave triangle mesh cropped around the ghost box, while the adapted object is represented as a dynamic rigid body initialized slightly above the fitted pose to avoid initial penetration. 
A short simulation with gravity, contact response, friction, and damping refines the final pose. 
We then apply a conservative grounding correction and a support-object-aware clamp to remove small residual gaps and prevent the object bottom from falling below the intended support surface. 
The clamp uses the predicted support object ID when available, avoiding correction to nearby but semantically incorrect geometry.

\section{Task Sets and Prompt Construction}
\label{app:tasks}

This section describes the task sets used in our evaluation and how prompts are constructed. 
We design two groups of tasks: a Common Placement Set for baseline-compatible placement evaluation, and a Fit-Critical Set for evaluating the complete ElasticFit under pose, contact, and local fitting constraints.

\begin{table}[h]
\centering
\caption{Task taxonomy used in the evaluation. 
All tasks include a base support condition specifying the intended placement surface; the listed types indicate the primary additional spatial or fit-critical constraint.}
\label{tab:task_taxonomy}
\resizebox{\linewidth}{!}{
\begin{tabular}{lll}
\toprule
\textbf{Task set} & \textbf{Subset / category} & \textbf{Primary constraint types} \\
\midrule
Common Placement 
& Distance 
& near, far \\
 
& Directional 
& left of, right of, in front of, behind \\
 
& Facing 
& face to \\
 
& Alignment 
& aligned with, parallel to \\
  
& Against wall 
& against wall \\
\midrule
Fit-Critical 
& Pose-and-contact 
& user-defined rotation, leaning, lying \\
 
& Gap-fitting 
& between objects, wall-side gap, under or inside space \\
\bottomrule
\end{tabular}
}
\end{table}

\subsection{Common Placement Set}

The Common Placement Set contains 60 tasks, denoted as S001--S060. 
It is used for cross-dataset validation and baseline comparison. 
All methods receive the same prompts, target object categories, and scene conditions, and all methods use fixed target assets so that the comparison focuses on placement quality rather than object generation or shape adaptation.

The set covers five primary relation categories, with 12 tasks assigned to each category:
\textit{distance}, \textit{directional}, \textit{facing}, \textit{alignment}, and \textit{against-wall} relations. 
These categories indicate the main relation being tested in each task, although individual prompts may include additional contextual constraints. 
Distance tasks include relations such as \textit{near} and \textit{far}. 
Directional tasks include \textit{left of}, \textit{right of}, \textit{in front of}, and \textit{behind}. 
Facing and alignment tasks evaluate object orientation with respect to reference objects, while against-wall tasks evaluate whether the object is placed near a wall-side region with a plausible orientation.

\paragraph{Prompt construction.} Common placement prompts follow templates such as: 
\begin{quote} 
\small 
Place a [object] on the [support object] near [reference object]. \\ 
Place a [object] on the [support object], to the left of [reference object]. \\ 
Place a [object] on the [support object], facing [reference object]. \\ 
Place a [object] on the [support object], aligned with [reference object]. \\ 
Place a [object] on the [support object] against the wall. 
\end{quote}

These tasks test whether a method can identify the intended placement region, ground the object in the scene, and satisfy semantic-spatial constraints without relying on object-shape adaptation.

\subsection{Fit-Critical Set}

The Fit-Critical Set contains 60 tasks, denoted as F001--F060. 
It is used to evaluate complete ElasticFit, including scene-conditioned object prior generation, 2D-to-3D lifting, and deformation-aware fitting. 
The set builds on the spatial-relation structure of the Common Placement Set, but adds pose, contact, and constrained-space requirements that require fit-aware insertion rather than rigid placement alone.

The Fit-Critical Set is divided into two subsets: pose-and-contact tasks and gap-fitting tasks.

\paragraph{Pose-and-contact tasks.}
The first 30 tasks, F001--F030, focus on explicit pose, orientation, and contact constraints. 
They extend common placement prompts with three additional relation types: \textit{user-defined rotation}, \textit{leaning}, and \textit{lying}. 
These tasks test whether a method can interpret pose-level instructions and produce physically plausible contact with support surfaces or nearby objects.

\paragraph{Prompt construction.}
Pose-and-contact prompts follow templates such as:
\begin{quote}
\small
Place a [object] on the [support object] rotated [angle] degrees. \\
Place a [object] on the [support object], leaning against [reference object]. \\
Place a [object] lying on the [support object].
\end{quote}

\paragraph{Gap-fitting tasks.}
The remaining 30 tasks, F031--F060, focus on constrained local spaces. 
They extend common placement prompts with three fit-critical relation types: \textit{between objects}, \textit{wall-side gap}, and \textit{under or inside space}. 
These tasks test whether a method can adapt the inserted object to available scene geometry, such as narrow gaps, side regions beside furniture, corners, and under- or inside-furniture spaces.

\paragraph{Prompt construction.}
Gap-fitting prompts follow templates such as:
\begin{quote}
\small
Place a [object] between [object A] and [object B] to fill the gap. \\
Place a [object] on the floor against the wall beside [reference object] to fill the side gap. \\
Place a [object] under the [furniture] to fill the available space. \\
Place a [object] inside the [furniture], parallel to it, to fill the available space.
\end{quote}

  
 
 
  
  
  
 
 
 

\section{Evaluation Metrics}
\label{app:metrics}

This section defines the evaluation metrics used in Sec.~\ref{experiments}. 
Unless otherwise stated, automatic metrics are computed from the final inserted object, annotated task relations, semantic instance annotations, and sampled scene surface points.

\subsection{Spatial Relation Success Rate}

For each task, the evaluator checks the annotated spatial relations between the inserted object and the referenced scene objects. 
For a prediction with relation set $\mathcal{R}$, the per-task spatial relation success is computed as
\begin{equation}
    \mathrm{SRS}_{\mathrm{task}}
    =
    \frac{1}{|\mathcal{R}_{\mathrm{eval}}|}
    \sum_{r \in \mathcal{R}_{\mathrm{eval}}}
    \mathbf{1}[r \text{ is satisfied}],
\end{equation}
where $\mathcal{R}_{\mathrm{eval}}$ denotes the subset of annotated relations that can be automatically evaluated from the available 3D geometry and metadata.

For distance relations, \textit{near} and \textit{far} are evaluated using the horizontal distance between the inserted object and the reference object, normalized by the reference object's horizontal bounding-box diagonal. 
For orientation relations, \textit{face to} is evaluated by comparing the predicted object forward direction with the direction toward the reference object, and \textit{parallel to}/\textit{aligned with} is evaluated by the angular difference between the object and reference directions. 
For support relations such as \textit{on top of}, the spatial check reuses the support success test described below.

View-dependent relations such as \textit{left of}, \textit{right of}, \textit{in front of}, and \textit{behind} can depend on the user-view frame, the reference object's intrinsic facing direction, or a dataset-specific coordinate convention. 
To avoid imposing an ambiguous frame convention in the automatic metric, SRS focuses on relations with stable 3D geometric definitions. 
Directional predicates are instead assessed through contextual plausibility scoring and qualitative analysis, where the rendered user view and original instruction are jointly available.

\subsection{Collision-Free Rate}

Collision-Free Rate measures whether the inserted object avoids unacceptable interpenetration with the existing scene. 
The scene mesh is sampled into a point set $\mathcal{P}$ with semantic object IDs. 
For each prediction, sampled scene points belonging to the annotated support object and wall-contact object are ignored when appropriate, since contact with these objects is expected.

Each remaining scene point is expressed in a yaw-aligned object frame for the horizontal coordinates, while its vertical coordinate is measured relative to the predicted object center. 
We count a scene sample as inside the predicted object box if it lies within the box extents after applying small horizontal and vertical boundary margins, which avoids treating boundary contact as collision. 
Let $N_{\mathrm{inside}}$ be the number of such inside scene samples. 
An insertion is considered collision-free if
\begin{equation}
    \mathrm{CFR}_{\mathrm{task}}
    =
    \mathbf{1}
    \left[
    N_{\mathrm{inside}} \leq \tau_{\mathrm{col}}
    \right],
\end{equation}
where $\tau_{\mathrm{col}}$ is a fixed tolerance for sparse sampling noise. 
The final Collision-Free Rate is the percentage of evaluated tasks whose collision-free indicator is $1$.

\subsection{Support Success Rate}

Support Success Rate measures whether the inserted object is physically supported by the annotated support surface. 
For each prediction, the evaluator collects scene points belonging to the annotated support object and lying within the predicted object footprint. 
The support height $z_{\mathrm{sup}}$ is estimated as the $95$th percentile of these footprint points, making the measurement robust to sparse or noisy samples. 
The object bottom height is computed from the predicted center and height as $z_{\mathrm{bottom}}=c_z-s_z/2$.

A placement is considered supported when the object bottom is sufficiently close to the estimated support height and the footprint has enough support coverage:
\begin{equation}
    \mathrm{SupR}_{\mathrm{task}}
    =
    \mathbf{1}
    \left[
    |z_{\mathrm{bottom}} - z_{\mathrm{sup}}| \leq \tau_{\mathrm{gap}}
    \ \wedge\
    C_{\mathrm{sup}} \geq \tau_{\mathrm{cover}}
    \right],
\end{equation}
where $C_{\mathrm{sup}}$ is the fraction of footprint support points lying within a small height band around $z_{\mathrm{sup}}$. 
In our implementation, we use $\tau_{\mathrm{gap}}=0.02\,\mathrm{m}$, a $0.02\,\mathrm{m}$ height band for coverage estimation, and $\tau_{\mathrm{cover}}=0.08$.

For CAD-style scenes or thin support surfaces, a bounding-box support variant can be used, where the object bottom is compared to the reference object's top bounding-box face and the XY footprint overlap is used as support coverage. 
The final Support Success Rate is the percentage of evaluated tasks whose support indicator is $1$.

\begin{figure}[h]
\centering
\fbox{
\begin{minipage}{0.94\linewidth}
\small
\ttfamily
You are an impartial evaluator for a 3D object placement task.\\[0.3em]

You will see multiple anonymized candidate placements for the SAME instruction.\\
The candidates are labeled only as Candidate A, Candidate B, Candidate C, and Candidate D.\\
The method names are hidden.\\[0.3em]

Evaluate ONLY the inserted target object with respect to the instruction and the visible scene context.\\
Ignore rendering quality aesthetics unless they make placement impossible to judge.\\[0.3em]

Instruction: \{prompt\}\\
Target object hint: \{target\_object\_hint\}\\[0.3em]

Task relations, if provided:\\
\{task\_relations\_text\}\\[0.3em]

Important rule:\\
Score each candidate using the 1--5 rubric.\\
You may use the other candidates only to calibrate obvious placement quality differences, but do NOT rank them directly.\\
Do NOT force all candidates to have different scores.\\
If multiple candidates are similarly plausible, they may receive the same score.\\[0.3em]

Do NOT penalize object identity/category mismatch, shape quality, or generation quality, unless the inserted object is missing, unrecognizable, or impossible to judge.\\
Evaluate only placement plausibility.\\
If a relation is visually ambiguous or hard to verify from the camera view, do NOT treat it as an issue.\\
Only penalize a relation when it is clearly violated.\\
When uncertain, give the benefit of the doubt.\\[0.3em]

Give each candidate a Contextual Plausibility Score from 1 to 5:\\
5 = Clearly satisfies the instruction and looks semantically natural.\\
4 = Satisfies the instruction overall; only minor placement awkwardness, but no clear violation.\\
3 = Main support/place is correct, but one important relation is clearly wrong or the placement is noticeably unnatural.\\
2 = Object is visible, but the support surface or multiple important relations are clearly wrong.\\
1 = Object is missing, unrecognizable, or placed in a clearly implausible/wrong region.
\end{minipage}
}
\caption{Prompt template used for Contextual Plausibility Score evaluation.}
\label{fig:cps_prompt}
\end{figure}

\subsection{Contextual Plausibility Score}

Contextual Plausibility Score (CPS) is computed using a grouped model-assisted evaluator.
For each task, anonymized screenshots of candidate placements are grouped and labeled only as Candidate A, B, C, and D; method names are hidden.
The evaluator receives the original instruction, a target-object hint, optional task relations, and the anonymized candidate images.
Candidates from the same task are scored together to calibrate placement quality differences, but the evaluator assigns an individual $1$--$5$ score to each candidate and is not forced to rank them or give different scores.
The evaluator is instructed to evaluate only placement plausibility and to give the candidate the benefit of the doubt when a relation is visually ambiguous.
Figure~\ref{fig:cps_prompt} shows the prompt template and scoring rubric. CPS has not been calibrated against human judgments and is used only as a complementary measure of placement plausibility, not of shape fidelity or deformation quality.



\subsection{Penetrating Vertex Ratio}

Penetrating Vertex Ratio (PVR) is used in the ablation analysis as a fine-grained diagnostic of object-scene interpenetration. 
For each fitted object, every mesh vertex queries its nearest scene sample, and a local signed distance is estimated using the corresponding scene normal. 
A vertex is counted as penetrating when the estimated signed distance is negative:
\begin{equation}
    \mathrm{PVR}
    =
    \frac{1}{|\mathcal{V}|}
    \sum_{i=1}^{|\mathcal{V}|}
    \mathbf{1}[d_i < 0],
\end{equation}
where $\mathcal{V}$ is the set of object vertices and $d_i$ is the estimated signed distance for vertex $v_i$. 
Lower values indicate that a smaller fraction of the object mesh penetrates the scene. 
Unlike the binary Collision-Free Rate, PVR captures finer differences between ablation variants.

\section{Baseline Adaptation Details}
\label{app:baselines}

We adapt LayoutGPT, Holodeck, and LayoutVLM to the single-object insertion setting used for the Common Placement Set. 
For all methods, the input scene remains fixed and only the target object is inserted. 
Each method receives the same scene, target object category, fixed target asset, and natural-language instruction. 
This controlled setting isolates semantic-spatial placement quality from differences in object generation or asset selection.

\begin{table}[htbp]
\centering
\caption{Summary of adapted baselines for the single-object insertion setting.}
\label{tab:baseline_adaptation_summary}
\begin{tabular}{p{0.22\linewidth}p{0.34\linewidth}p{0.34\linewidth}}
\toprule
\textbf{Method} & \textbf{Spatial semantics} & \textbf{Physical plausibility} \\
\midrule
LayoutGPT &
Directly predicts target position and yaw from a CSS-style scene description. &
No explicit collision or support refinement; placements are only clamped to scene bounds. \\[0.25em]

Holodeck &
Generates Holodeck-style spatial constraints and searches over discrete placement candidates. &
Uses AABB-level collision, support, wall, and relation checks during candidate scoring. \\[0.25em]

LayoutVLM &
Initializes object pose from vision-language reasoning and semantic constraints. &
Uses bounding-box-level differentiable optimization for feasible placement refinement. \\[0.25em]

ElasticFit &
Uses view-aligned structured grounding with support validation and local scene cues. &
Uses geometry-aware local mesh grounding and physics-guided placement refinement. \\
\bottomrule
\end{tabular}
\end{table}

\paragraph{LayoutGPT.}
LayoutGPT is adapted as a text-only layout prediction baseline for single-object insertion. 
Following the original LayoutGPT formulation, the scene is represented as a CSS-style structured layout, where each existing object is described by its semantic name, caption, 3D bounding-box size, center position, and yaw orientation. 
For each task, the input prompt provides the fixed scene bounds, a list of selected context objects, the target object category and fixed asset size, and the user instruction. 
The model is asked to output exactly one CSS-style block for the inserted target object:
\begin{equation}
    o_T = (\ell_T, w_T, h_T, x_T, y_T, z_T, \theta_T),
\end{equation}
where $d_T=(\ell_T, w_T, h_T)$ denotes the fixed asset dimensions, $p_T=(x_T,y_T,z_T)$ is the predicted object center, and $\theta_T$ is the predicted yaw orientation. 
The existing scene objects are kept unchanged, and no collision, support, scaling, deformation, pitch, or roll refinement is applied after prediction. 
Predicted positions are only clamped to the scene bounds before evaluation.

\begin{table}[htbp]
\centering
\caption{Prompt structure for the adapted LayoutGPT baseline.}
\label{tab:layoutgpt_prompt_structure}
\resizebox{\linewidth}{!}{
\begin{tabular}{ll}
\toprule
\textbf{Prompt block} & \textbf{Purpose} \\
\midrule
Task instruction & Specifies single-object insertion into a fixed existing 3D scene. \\
Coordinate system & Defines \texttt{left}, \texttt{depth}, \texttt{top}, and yaw orientation. \\
Scene bounds & Provides the valid 3D range of the scene in centimeters. \\
Output template & Enforces one CSS-style block for the target object. \\
Rules & Predicts only position and yaw while keeping the target asset fixed. \\
Scene objects & Lists context objects with captions, size, position, and orientation. \\
Target object size & Provides fixed length, width, and height of the inserted asset. \\
User instruction & Provides the natural-language placement request. \\
\bottomrule
\end{tabular}
}
\end{table}

\paragraph{Holodeck.}
Holodeck is adapted as a constraint-generation and search-based placement baseline while preserving the main design of the original method. 
The original Holodeck uses an LLM to generate spatial relational constraints between objects, rather than directly predicting absolute bounding-box values, and then searches for layouts that satisfy these constraints while enforcing hard constraints such as collision avoidance and room-boundary validity~\cite{yang2024holodeck}. 
In the adapted single-object insertion setting, the existing scene is kept fixed and the same principle is applied only to the inserted target object. 
This preserves Holodeck's core design choice: language-based relation generation followed by solver-based placement search.

First, a language model receives the user instruction, target object description, target asset size, and a list of pre-existing scene objects described by ID, semantic label, center, size, yaw, and coarse location. 
It outputs a Holodeck-style constraint line for \texttt{object-target}, using relations such as \textit{near}, \textit{left of}, \textit{on top of}, \textit{face to}, \textit{align}, and \textit{against wall}. 
The output is parsed into relation-anchor pairs, and invalid relations or anchors are discarded.

Second, an AABB-based solver generates discrete candidate placements around the referenced anchors. 
Directional relations sample candidates around the anchor according to its yaw, \textit{on top of} uses dense sampling on the anchor's top AABB surface, and \textit{against wall} samples candidates along semantic wall AABBs. 
Candidates are ranked using relation scores, floor-bound checks, AABB-level collision checks, and support validation. 
The target asset remains fixed, and no object generation, deformation, pitch/roll estimation, local mesh grounding, or physics-based settling is applied.

\paragraph{LayoutVLM.}
We adapt LayoutVLM as a vision-language-guided placement baseline for single-object insertion. 
All existing scene objects are kept fixed, and the method predicts and refines only the inserted target object. 
For each task, the VLM receives three visual inputs: a top-down sliced scene plan with grid lines and object IDs, a perspective rendering of the current scene, and a canonical rendering of the target asset. 
It also receives a structured scene program describing the fixed scene objects with semantic labels, bounding-box sizes, centers, and yaw orientations.

The VLM is prompted to output a LayoutVLM-style Python program containing an initial target pose and optional solver-style spatial relations, such as distance constraints, \texttt{on\_top\_of}, \texttt{point\_towards}, \texttt{align\_with}, and \texttt{against\_wall}. 
We parse the predicted target position, yaw, and relations from this program. 
When an explicit relation in the instruction is omitted, such as \emph{near}, \emph{far}, \emph{on top of}, or \emph{facing}, we recover the corresponding prompt-required constraint from the instruction. 
We then apply a self-consistency filter that removes VLM-predicted relations incompatible with the predicted initial pose, while preserving prompt-required relations.

Given the VLM-predicted initial pose $(p_0,\theta_0)$ and the parsed constraint set $\mathcal{C}$, the adapted LayoutVLM solver optimizes only the target center $p$ and yaw $\theta$:
\[
\min_{p,\theta}
\lambda_{\mathrm{rel}} L_{\mathrm{rel}}(p,\theta;\mathcal{C})
+ \lambda_{\mathrm{sup}} L_{\mathrm{sup}}(p,\theta)
+ \lambda_{\mathrm{col}} L_{\mathrm{AABB}}(p,\theta)
+ \lambda_{\mathrm{bd}} L_{\mathrm{bd}}(p,\theta)
+ \lambda_{\mathrm{init}}\|p-p_0\|_2 .
\]
Here, $L_{\mathrm{rel}}$ encodes the parsed spatial relations, $L_{\mathrm{sup}}$ enforces support containment for \texttt{on\_top\_of} relations, $L_{\mathrm{AABB}}$ penalizes bounding-box intersections with fixed scene objects, $L_{\mathrm{bd}}$ keeps the target inside the scene bounds, and the initialization term preserves the VLM-predicted pose.

The target asset size is kept fixed throughout this process. 
Physical feasibility is approximated using bounding-box-level collision, support, and boundary checks. 
No object generation, mesh-level local grounding, elastic deformation, pitch/roll adaptation, or physics-based settling is applied. 
Thus, this baseline evaluates vision-language-guided pose prediction and bounding-box-level optimization without ElasticFit's scene-conditioned generation, and deformation-aware fitting.

\paragraph{ElasticFit placement-only setting.}
For the Common Placement Set baseline comparison, ElasticFit is evaluated in a placement-only setting. 
Scene-conditioned object generation and deformation-aware fitting are disabled, and a fixed target asset is used for each task. 
The method retains view-aligned VLM grounding, target validation through the ID buffer, and geometry- and physics-aware placement refinement. 
This setting isolates ElasticFit's placement and grounding quality from the additional cost of object generation and 3D reconstruction.

\begin{table}[h]
\centering
\caption{Average runtime comparison on the Common Placement Set.}
\label{tab:runtime_comparison}
\begin{tabular}{lcl}
\toprule
\textbf{Method} & \textbf{Avg. total time (s)} & \textbf{Notes} \\
\midrule
LayoutGPT& 42.97 & End-to-end runtime \\
Holodeck & 16.71 & End-to-end runtime \\
LayoutVLM & 100.03 & End-to-end runtime \\
ElasticFit & 11.53 & Placement-only; 10.50 s VLM + 1.03 s fitting \\
\bottomrule
\end{tabular}
\end{table}

\paragraph{Runtime and Computational Cost.}
\label{app:runtime}

Table~\ref{tab:runtime_comparison} reports the average runtime in the adapted single-object insertion setup. 
In the placement-only setting, ElasticFit takes 11.53 seconds per task on average, including 10.50 seconds for VLM reasoning and 1.03 seconds for local fitting and physics refinement. 
ElasticFit is faster than all three adapted baselines in this setting, with most runtime spent on VLM reasoning. 
Since runtime depends on implementation details, hardware, API latency, and retry behavior, these numbers should be interpreted as measurements for our experimental setup rather than universal speed comparisons.

\paragraph{Compute Resources.}
All local experiments were run on a workstation with an NVIDIA GeForce RTX 4070 Ti SUPER GPU with 16 GB dedicated GPU memory and approximately 64 GB system RAM. 
CPU preprocessing, scene rendering, geometric fitting, physics simulation, and evaluation scripts were executed on the same local workstation. 
VLM reasoning and model-assisted evaluation were performed through remote API calls, while image generation and 2D-to-3D reconstruction were executed locally when required. 
No model training or fine-tuning was performed.

The complete ElasticFit method additionally requires scene-conditioned image generation and 2D-to-3D reconstruction, so its runtime is higher and depends on generation, reconstruction, API latency, and retry behavior. 
The reported experiments include the 60-task Common Placement Set, the 60-task Fit-Critical Set, and the 30-task gap-fitting ablation subset. 
Total compute is estimated from logged per-task runtimes rather than from a cluster-level accounting system, and excludes preliminary debugging runs, failed API calls, and exploratory parameter tuning.

\section{Additional Qualitative Examples}
\label{app:additional_qualitative}

This section provides additional qualitative examples for the Common Placement Set and the Fit-Critical Set.

\subsection{Additional Common Placement Results}

\begin{figure}[t]
    \centering
    \includegraphics[width=\linewidth]{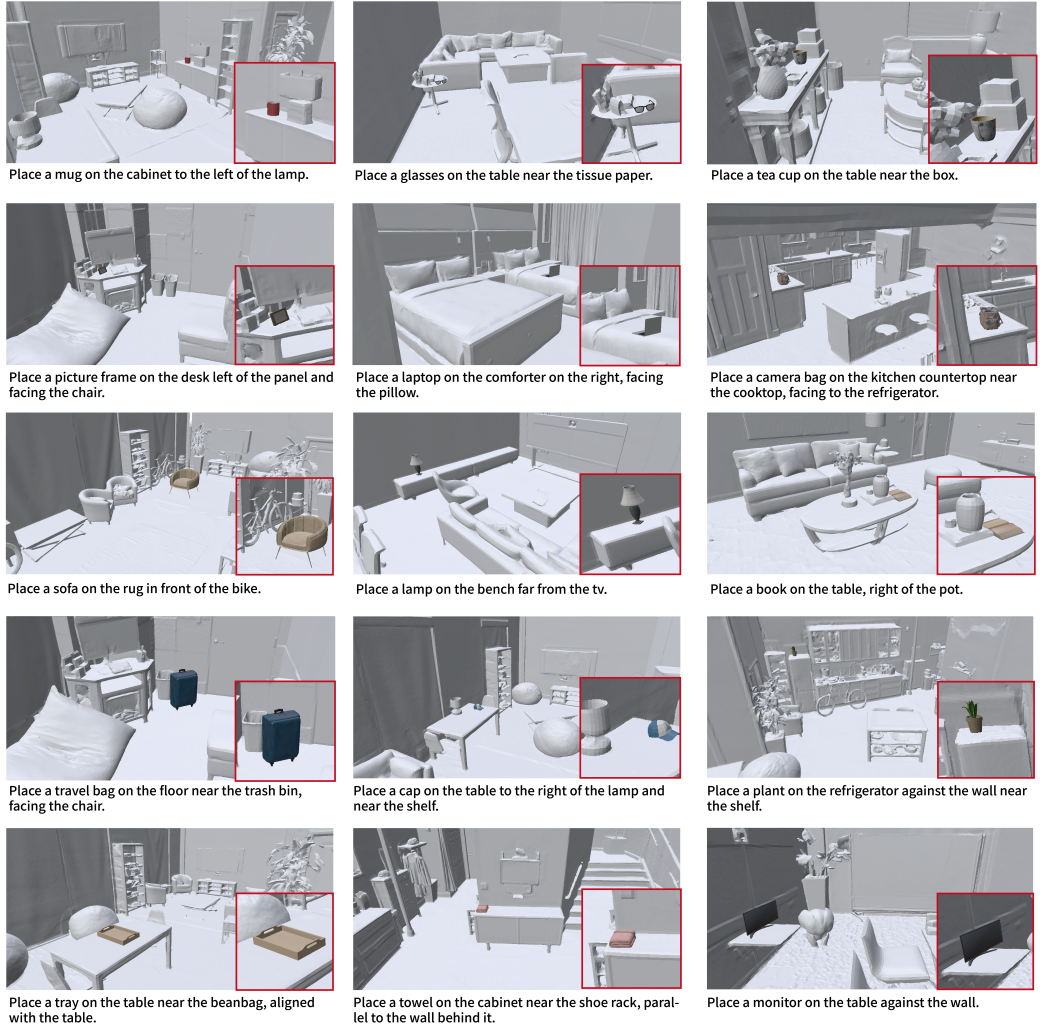}
    \caption{
    Additional qualitative results on the Common Placement Set.
    Each example shows the final fitted placement produced by ElasticFit, with red boxes highlighting the inserted object.
    The examples include diverse spatial relations such as left/right, near/far, facing, alignment, and against-wall placement.
    }
    \label{fig:additional_common_placement}
\end{figure}

Figure~\ref{fig:additional_common_placement} shows additional qualitative results on the Common Placement Set. 
These examples cover diverse spatial relations and illustrate that ElasticFit produces placements that are spatially grounded, supported by the intended scene geometry, and visually consistent with the input instruction.

\subsection{Additional Fit-Critical Results}
\label{app:additional_fit_critical}

\begin{figure}[!htbp]
    \centering
    \includegraphics[width=\linewidth]{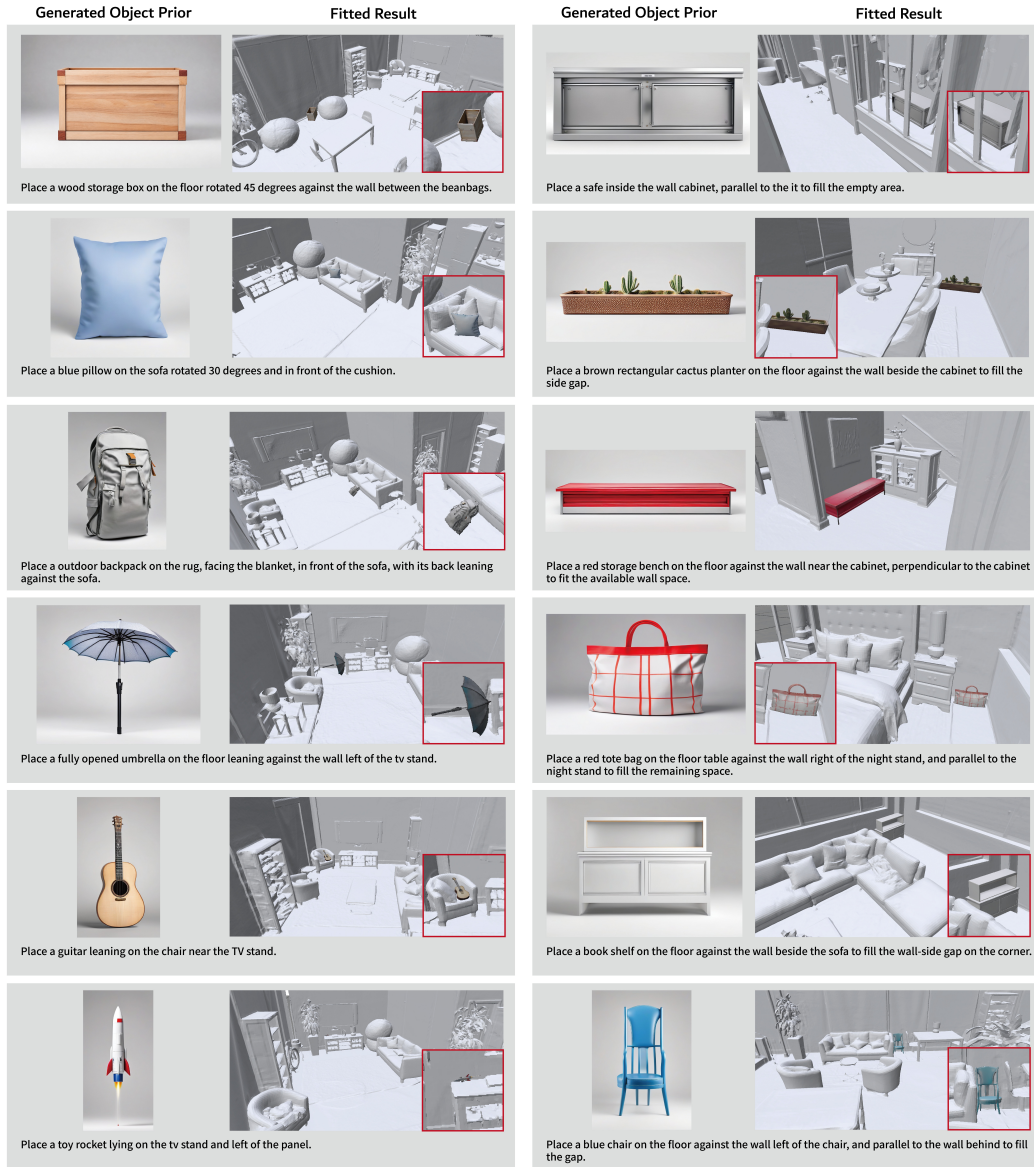}
    \caption{
    Additional qualitative results on the Fit-Critical Set.
    The examples show final fitted insertions produced by ElasticFit, including pose-and-contact cases and gap-fitting cases in constrained local scene regions.
    }
    \label{fig:additional_fit_results}
\end{figure}

Figure~\ref{fig:additional_fit_results} shows additional qualitative results on the Fit-Critical Set.
These examples illustrate final fitted insertions produced by ElasticFit under pose-and-contact constraints and gap-fitting constraints in constrained local scene regions.

\section{Asset Licenses}
\label{app:asset_licenses}

Table~\ref{tab:asset_licenses} lists the third-party 3D assets used in our experiments.
All assets were obtained from Sketchfab \cite{sketchfab} under the Creative Commons Attribution 4.0 (CC BY 4.0) license.
We use these assets in accordance with their license terms and do not redistribute the original third-party asset files.

\begin{table}[!htbp]
\centering
\caption{Third-party 3D assets used in the experiments.}
\label{tab:asset_licenses}
\small
\setlength{\tabcolsep}{10pt}
\begin{tabular}{lll}
\toprule
\textbf{Assets}  &  &   \\
\midrule
Alarm clock & Book & Chair \\
Backpack & Bowl & Coffee machine \\
Bike & Camera bag & Fan \\
Camera & Cap & Glasses \\
Lamp & Laptop & Monitor \\
Mug & Picture frame & Pillow \\
Plant & Salad plate & Small oven \\
Sofa & Speaker & Tea cup \\
Teddy bear & Towel & Toy car \\
Toy rabbit & Travel bag & Tray \\
Vase & & \\
\bottomrule
\end{tabular}
\end{table}




\end{document}